\documentclass[10pt,twocolumn,letterpaper]{article}

\usepackage[pagenumbers]{cvpr} 

\definecolor{cvprblue}{rgb}{0.21,0.49,0.74}
\usepackage[accsupp]{axessibility}  
\usepackage[pagebackref,breaklinks,colorlinks,allcolors=cvprblue]{hyperref}
\usepackage{multirow, multicol}
\usepackage{wrapfig}
\usepackage[table]{xcolor}
\usepackage{graphicx}
\usepackage{gradient-text}
\usepackage[most]{tcolorbox}
\usepackage{balance}
\usepackage{pifont}
\definecolor{oursbg}{HTML}{EEDEFF}
\definecolor{sbred}{HTML}{DB6057}
\definecolor{sbgreen}{HTML}{92DB58}
\definecolor{sbblue}{HTML}{5771DB}
\definecolor{sbpurple}{HTML}{A157DB}

\definecolor{firstbg}{HTML}{E6CFFF}
\definecolor{secondbg}{HTML}{F3E8FF}
\definecolor{thirdbg}{HTML}{FAF4FF}
\definecolor{positivebg}{HTML}{E3F6D2}
\definecolor{negativebg}{HTML}{FBD4CE}
\definecolor{oursbar}{HTML}{5771db}
\definecolor{v2sbar}{HTML}{db6057}

\newcommand{\cc}{\cellcolor{oursbg}}
\newcommand{\first}{\cellcolor{firstbg}}
\newcommand{\second}{\cellcolor{secondbg}}
\newcommand{\third}{\cellcolor{thirdbg}}
\newcommand{\posc}{\cellcolor{positivebg}}
\newcommand{\negc}{\cellcolor{negativebg}}
\newcommand{\titleLogo}{%
  \raisebox{-0.35\height}{\includegraphics[width=0.07\textwidth]{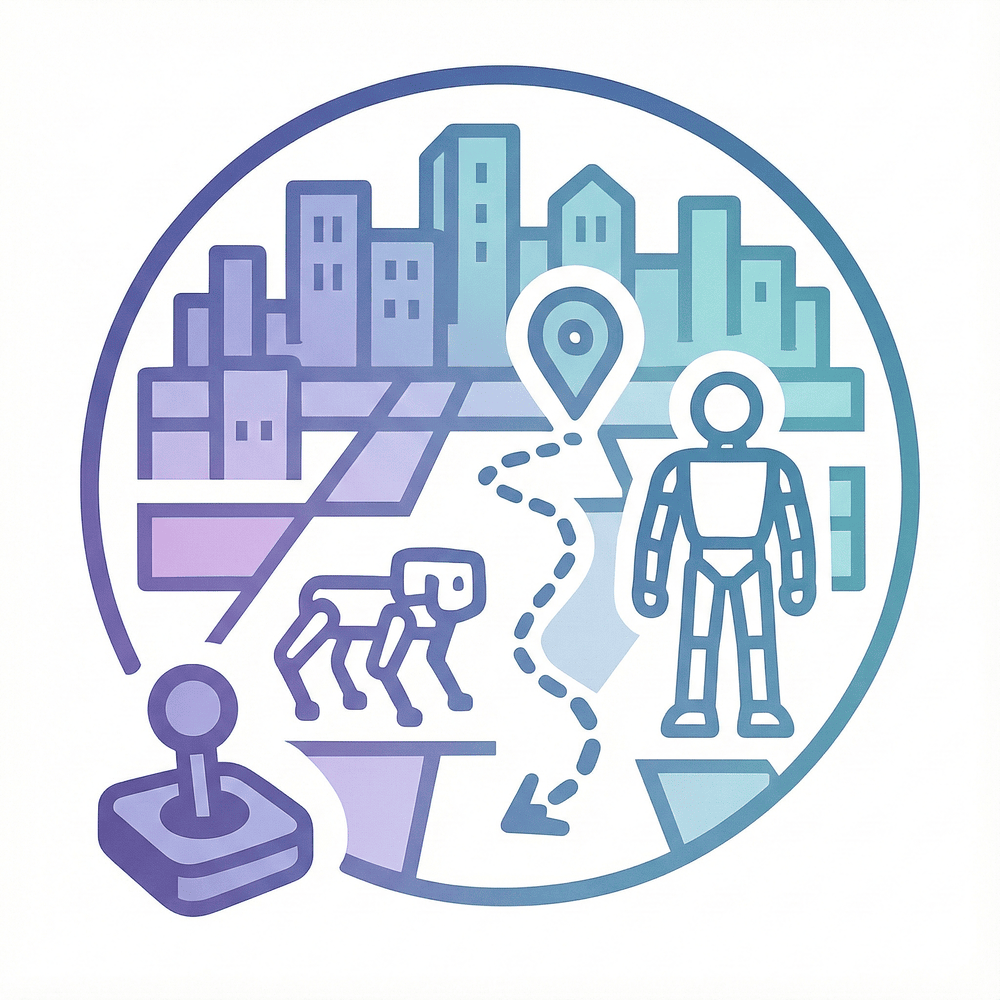}}%
}
\newcommand{\wanderland}{\textsc{\gradientRGB{Wanderland}{16,185,129}{76,29,149}}}

\newtcolorbox{redbox}{
  colback=gray!10!white,
  colframe=red!40!white,
  boxrule=0.6pt,
  arc=2mm,
  left=6pt, right=6pt, top=3pt, bottom=3pt,
  before skip=10pt, after skip=10pt
}
\newtcolorbox{greenbox}{
  colback=gray!10!white,
  colframe=green!40!white,
  boxrule=0.6pt,
  arc=2mm,
  left=6pt, right=6pt, top=3pt, bottom=3pt,
  before skip=10pt, after skip=10pt
}
\newtcolorbox{bluebox}{
  colback=gray!10!white,
  colframe=blue!40!white,
  boxrule=0.6pt,
  arc=2mm,
  left=6pt, right=6pt, top=3pt, bottom=3pt,
  before skip=10pt, after skip=10pt
}

\def\paperID{xxxx} 
\def\confName{CVPR}
\def\confYear{2026}

\title{\titleLogo \textsc{\gradientRGB{Wanderland}{16,185,129}{76,29,149}}: \\ Geometrically Grounded Simulation for Open-World Embodied AI}

\author{Xinhao Liu\textsuperscript{*,1} \quad Jiaqi Li\textsuperscript{*,1} \quad Youming Deng\textsuperscript{2} \quad Ruxin Chen\textsuperscript{1} \quad Yingjia Zhang\textsuperscript{1} \\ Yifei Ma\textsuperscript{1} \quad Li Guo\textsuperscript{1} \quad Yiming Li\textsuperscript{1} \quad Jing Zhang\textsuperscript{\ding{41},1} \quad Chen Feng\textsuperscript{\ding{41},1} \\ New York University\textsuperscript{1} \quad \quad Cornell University\textsuperscript{2} \\
{\small \bf \url{https://ai4ce.github.io/wanderland/}}
}

\begin{document}
\twocolumn[{
    \renewcommand\twocolumn[1][]{#1}
    \maketitle
    \vspace{-2em}
    \begin{center}
        \centering
        \includegraphics[width=0.95\textwidth]{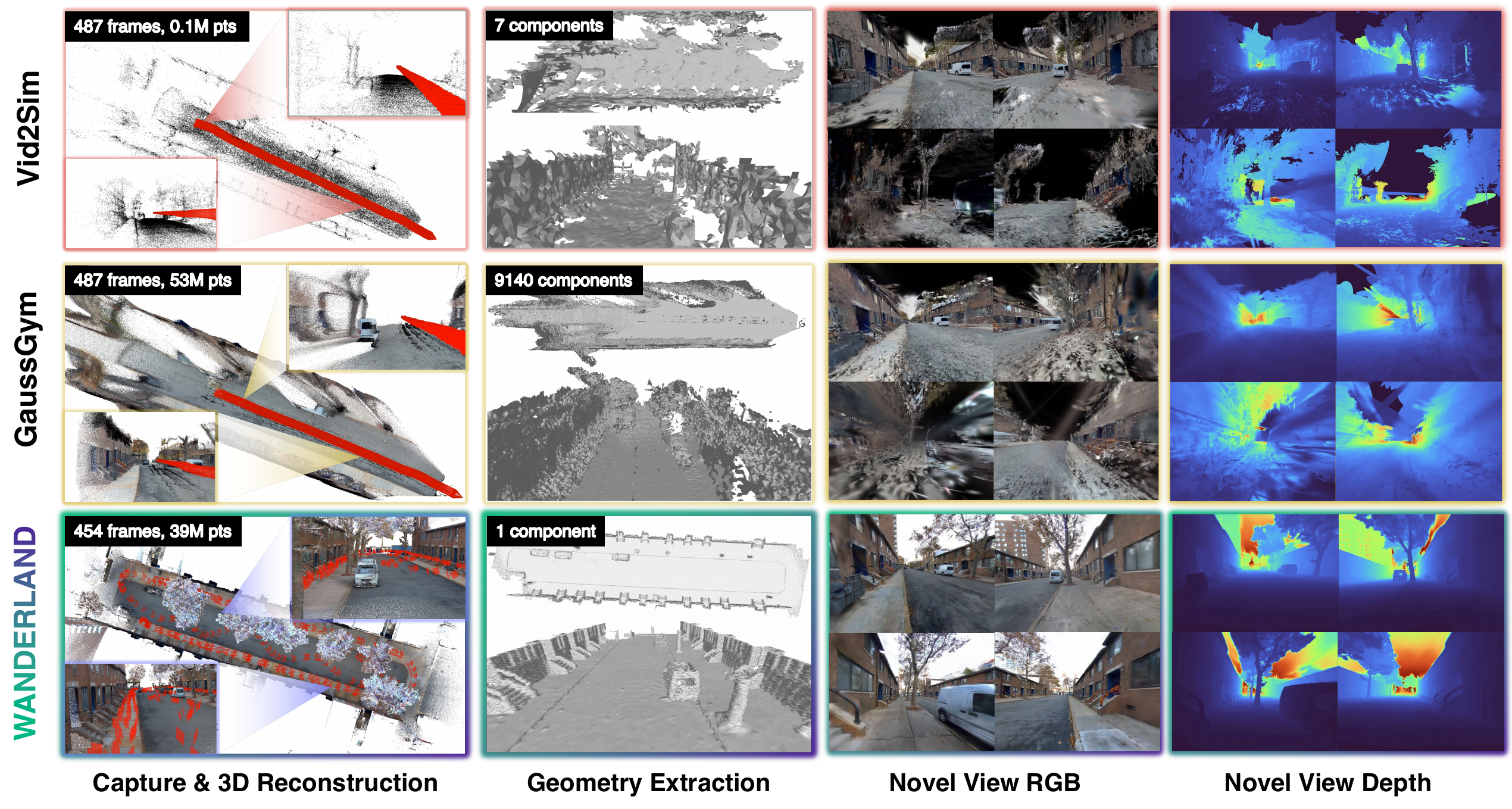}
        \vspace{-2mm}
        \captionof{figure}{\textbf{Do video-3DGS frameworks provide geometrically grounded and photorealistic simulation?} We demonstrate that building such simulations from casually captured touring videos often fails due to limited view diversity, inaccurate 3D reconstruction, unreliable geometry extraction, and degraded novel-view extrapolation. We propose the \wanderland{} framework that features multi-sensor diverse-view capture, reliable reconstruction, accurate metric-scale geometry, and robust view synthesis.}
        \label{fig:teaser}
    \end{center}
}]
{
  \renewcommand{\thefootnote}%
    {\fnsymbol{footnote}}
  \footnotetext[1]{Equal contribution, random order.}
  \footnotetext{\thanks{\ding{41} Corresponding authors, \texttt{\{jz6676,cfeng\}@nyu.edu}}}
}
\begin{abstract}
Reproducible closed-loop evaluation remains a major bottleneck in Embodied AI such as visual navigation. A promising path forward is high-fidelity simulation that combines photorealistic sensor rendering with geometrically grounded interaction in complex, open-world urban environments.
Although recent video-3DGS methods ease open-world scene capturing, they are still unsuitable for benchmarking due to large visual and geometric sim-to-real gaps. To address these challenges, we introduce Wanderland, a real-to-sim framework that features multi-sensor capture, reliable reconstruction, accurate geometry, and robust view synthesis. Using this pipeline, we curate a diverse dataset of indoor-outdoor urban scenes and systematically demonstrate how image-only pipelines scale poorly, how geometry quality impacts novel view synthesis, and how all of these adversely affect navigation policy learning and evaluation reliability. Beyond serving as a trusted testbed for embodied navigation, Wanderland's rich raw sensor data further allows benchmarking of 3D reconstruction and novel view synthesis models. Our work establishes a new foundation for reproducible research in open-world embodied AI.
\end{abstract}
\vspace{-5mm}    
\section{Introduction}
\label{sec:intro}
\vspace{-2mm}
Embodied AI research has advanced significantly in recent years, driven by high-performance simulation platforms~\cite{savva2019habitat,shen2021igibson,kolve2017ai2,gan2020threedworld,xiang2020sapien,li2024behavior} and datasets~\cite{ramakrishnan2021habitat,replica19arxiv,chang2017matterport3d,deitke2020robothor,dai2017scannet,xia2018gibson}. They have played a central role in this progress by standardizing tasks, metrics, and providing closed-loop evaluation. With the emergence of foundation models for vision, language, and decision making~\cite{bai2025qwen2,liu2024improved,beyer2024paligemma,liu2025nvila}, it is natural to extend embodied AI beyond domestic interiors to open-world settings. Applications in embodied navigation include last-mile delivery~\cite{wasserman2023last}, campus-scale wayfinding~\cite{lee2024learning,carlevaris2016university,liang2025gnd}, and service robots that must traverse lobbies, hallways, plazas, and sidewalks~\cite{liu2025citywalker,sorokin2022learning,peng2025data}. These use cases require simulative environments with large spatial extents, mixed indoor-outdoor coverage, high-fidelity sensor simulation, and reliable physics for interaction. They also require datasets with diverse scene coverage, such as sidewalks, grocery stores, or subway stations. \textit{How can we build a high-fidelity real-to-sim testbed to evaluate and benchmark open-world navigation systems?}

This demand poses significant challenges to classic datasets and their curation pipelines using RGB-D sensors. These challenges begin at capture: prevalent RGB-D sensors fail outdoors due to sunlight interference and limited range, and tripod-based methods are inefficient for large-scale areas. During reconstruction, pose estimation via RGB-D fusion is prone to drift and failure in large-scale, low-texture, and less-structured environments. Finally, simulation fidelity is further affected by low-quality texture meshes reconstructed without enough spatial resolution, or with artifacts like non-watertight and fragmented geometry, especially outdoors. Consequently, these curation pipelines cannot reliably support large-scale, high-fidelity simulations in open-world environments.

\begin{table*}[t]
    \caption{\textbf{Real-to-sim datasets for embodied navigation}. Classic datasets are mostly from dedicated tripod-based capture, providing metric scale environments, and representing scenes as textured meshes, but they are \textit{limited to indoor scenes}. 3DGS datasets can include outdoor scenes, but they suffer from \textit{inaccurate 3D reconstruction} and \textit{non-metric-scale} environments. $^*$: Most scenes are from ARKitScenes~\cite{dehghan2021arkitscenes} and ANYmal GrandTour datasets~\cite{Frey2025Boxi}. \raisebox{-0.15ex}{\includegraphics[height=1em]{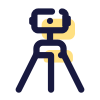}}: Tripod-based capture. \raisebox{-0.15ex}{\includegraphics[height=1em]{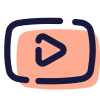}}: Casual online videos. \raisebox{-0.15ex}{\includegraphics[height=1em]{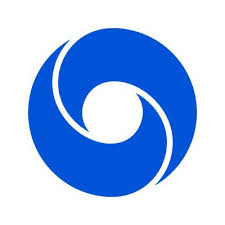}}: Generated videos. \raisebox{-0.15ex}{\includegraphics[height=1em]{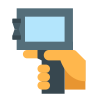}}: Mobile 3D scanner.}
    \vspace{-2mm}
    \label{tab:datasets}
    \centering
    \resizebox{\linewidth}{!}{
    \begin{tabular}{lccccccccc}
        \toprule
         \textbf{Dataset} & \textbf{Source} & \textbf{Capture} & \textbf{Scene} & \textbf{Metric Scale} & \textbf{\#Scenes} & \textbf{\#Frames} & \textbf{Scene Size (m$^2$)}  & \textbf{Geometry Source} & \textbf{Reconstruction}\\
         \midrule
         \multicolumn{10}{l}{\cellcolor[HTML]{E8E8E8} \textit{Classic Datasets}} \\
         Matterport3D~\cite{chang2017matterport3d} & \raisebox{-0.15ex}{\includegraphics[height=1.1em]{icons/tripod.png}} & RGB-D & Indoor & \checkmark & 90 & 11K & 102K & Depth & Matterport\\
         ScanNet~\cite{dai2017scannet} & \raisebox{-0.15ex}{\includegraphics[height=1.1em]{icons/handheld.png}} & RGB-D, IMU & Indoor & \checkmark & 1,513 & 2.5M & 40K & Depth & BundleFusion~\cite{dai2017bundlefusion} \\
         Gibson~\cite{xia2018gibson} & \raisebox{-0.15ex}{\includegraphics[height=1.1em]{icons/tripod.png}} & RGB-D & Indoor & \checkmark & 572 & --- & 218K & Depth & Proprietary \\
         HM3D~\cite{ramakrishnan2021habitat} & \raisebox{-0.15ex}{\includegraphics[height=1.1em]{icons/tripod.png}} & RGB-D & Indoor & \checkmark & 1,000 & --- & 365K & Depth & Matterport \\
         \midrule
         \multicolumn{10}{l}{\cellcolor[HTML]{E8E8E8} \textit{3DGS Datasets}} \\
         Vid2Sim~\cite{xie2025vid2sim} & \raisebox{-0.15ex}{\includegraphics[height=1.1em]{icons/youtube.png}} & RGB & Outdoor  & $\times$ & 30 & 13.5K & --- & 3DGS Opacity & GLOMAP~\cite{pan2024global} \\
         GaussGym~\cite{escontrela2025gaussgym} & \raisebox{-0.15ex}{\includegraphics[height=1.1em]{icons/youtube.png}\includegraphics[height=1.1em]{icons/deepmind.png}} & RGB & Mixed & $\times$ & 2,500$^*$ & --- & --- & Point Map & VGGT~\cite{wang2025vggt} \\
         \midrule
         \cc \bf \wanderland{} (ours) & \cc \bf \raisebox{-0.5ex}{\includegraphics[height=1.1em]{icons/handheld.png}} & \cc \bf LiDAR, IMU, RGB & \cc \bf Mixed & \cc \bf \checkmark & \cc \bf 530 & \cc \bf 420K & \cc \bf 3.8M & \cc \bf LiDAR & \cc \bf LIV SLAM\\
         \bottomrule
    \end{tabular}
    }
    \vspace{-2mm}
\end{table*}

To overcome the limitations of indoor-only datasets, recent work has explored using online videos as an alternative data source for building open-world environments. Methods based on 3D Gaussian Splatting (3DGS)~\cite{kerbl20233d} have shown promise in creating visually compelling simulations from casual videos and enabling navigation policy training via reinforcement learning~\cite{adamkiewicz2022vision,lei2025gaussnav,chen2025splat,chhablani2025embodiedsplat}. However, these video-based approaches face fundamental challenges that prevent their use as standardized benchmarks: 

First, they suffer from \textbf{inaccurate 3D reconstruction} due to reliance on pure RGB-based methods like SfM~\cite{schonberger2016structure,pan2024global} or deep reconstruction models~\cite{leroy2024grounding,wang2025vggt}, yielding non-metric camera poses and depth estimations. Second, they produce \textbf{unreliable geometry} for physical interaction simulation, with collision meshes extracted from 3DGS opacity fields being fragmented and metrically ungrounded~\cite{guedon2024sugar,yu2024gaussian}. Finally, they exhibit severe \textbf{extrapolated view degradation} due to uniform video trajectories, causing rendering quality to sharply decline for viewpoints beyond the capture path~\cite{han2025extrapolated}. Consequently, while suitable for training, these environments lack the geometric grounding required for reproducible benchmarking of embodied navigation systems.

To bridge this gap demonstrated in \cref{fig:teaser}, we present \wanderland, a robust real-to-sim framework featuring multi-sensor diverse-view capture, reliable reconstruction, accurate metric-scale geometry, and robust view synthesis. Our pipeline begins with data capturing with a handheld multi-sensor 3D scanner (\cref{fig:metacam}). This setup enables dense, multi-view capture across large-scale indoor and outdoor scenes, providing rich and diverse visual coverage. Our reconstruction leverages a LiDAR-inertial-visual (LIV) SLAM system, which fuses these complementary sensor streams to produce globally consistent, metric-scale point clouds and highly accurate camera poses. 

This reliable geometric foundation directly addresses the limitations of video-only pipelines, ensuring metric accuracy and completeness. We then utilize this precise geometry to initialize high-quality 3DGS models and extract clean, reliable collision meshes. Finally, the framework integrates the 3DGS model and geometrically grounded mesh into Isaac Sim, creating a unified environment that supports both photorealistic rendering and geometrically grounded interaction. As shown in \cref{tab:datasets}, our approach achieves the best of both worlds: efficient open-world capture and geometrically accurate reconstruction.

Leveraging this framework, we introduce the \wanderland{} dataset, which serves not only as a geometrically grounded and photorealistic simulation environment, but also as a critical diagnostic tool for understanding the limitations of existing approaches. We use the dataset to systematically demonstrate how and why vision-only reconstruction methods fail to provide reliable geometric groundings for embodied AI, particularly in open-world settings where metric accuracy and consistent novel view synthesis are essential. 

Beyond identifying these issues, \wanderland{} establishes a trusted testbed for benchmarking embodied navigation tasks like image-goal and vision-language navigation under closed-loop control, enabling reliable evaluation in environments that are both visually realistic and geometrically consistent. Furthermore, our rich raw sensor data provides a valuable benchmark for core 3D vision tasks, supporting comprehensive evaluation of 3D reconstruction and novel view synthesis methods. This multi-faceted design enables ablation studies on simulation construction and offers key insights towards advancing large-scale 3D vision and embodied AI research.

We summarize our contributions as follows:
\begin{enumerate}
    \item We identify fundamental limitations of video-3DGS pipelines for embodied AI and introduce a new real-to-sim framework that overcomes these challenges.
    \item We introduce \wanderland{} framework and dataset for systematic analysis of key factors in simulation construction. It also serves as a testbed for embodied navigation.
    \item We highlight that visual-only reconstructions remain significantly less accurate than LiDAR-enhanced ones, resulting in unreliable simulation for embodied AI.
    \item We provide rich raw sensor data that enables benchmark evaluation for 3D reconstruction and novel view synthesis, supporting method evaluation and ablation studies.
\end{enumerate}

\vspace{-2mm}
\section{Related Work}
\label{sec:related}
\textbf{Real-to-sim datasets for embodied navigation}. Classic embodied navigation datasets such as Matterport3D~\cite{chang2017matterport3d}, ScanNet~\cite{dai2017scannet}, and HM3D~\cite{ramakrishnan2021habitat} provide metric-scale environments through dedicated RGB-D capture and mesh-based reconstruction, but remain limited to indoor settings. In contrast, our \wanderland{} dataset employs handheld multi-sensor capture and 3DGS representation, enabling high-fidelity sensor simulation across diverse indoor-outdoor environments while maintaining metric accuracy.

Recent video-based approaches expand to open-world settings using alternative data sources. Vid2Sim~\cite{xie2025vid2sim} reconstructs 30 outdoor scenes from 13.5K YouTube frames using GLOMAP for SfM initialization and 3DGS opacity for geometry. GaussGym~\cite{escontrela2025gaussgym} aggregates 2,500 mixed scenes from online and generated videos, employing VGGT~\cite{wang2025vggt} and NKSR~\cite{huang2023neural} for reconstruction. Compared to these vision-only methods that rely on visual inference from casual video sources, \wanderland{} utilizes multi-modal sensing and LIV-SLAM reconstruction, providing geometrically accurate foundations for physical interaction while achieving visual fidelity through 3DGS rendering.

\noindent\textbf{Open-world 3D reconstruction}. Robust 3D reconstruction for open-world environments remains challenging due to the limitations of existing capture methodologies. Many outdoor scene datasets~\cite{ling2024dl3dv,tung2024megascenes} rely on SfM methods like COLMAP~\cite{schonberger2016structure} for sparse reconstruction. Due to its vision-only nature, SfM methods can't provide metric scale camera poses and are prone to errors. In contrast to the image set input in SfM, SLAM algorithms take sequential input. Visual(-inertial) SLAM systems~\cite{mur2015orb,forster2014svo,sumikura2019openvslam,teed2023deep,lipson2024deep,qin2018vins} improve tracking but still lack absolute scale and global consistency over long trajectories. LiDAR(-inertial) SLAM~\cite{shan2018lego,shan2020lio,xu2022fast,chen2023deepmapping2,guadagnino2025kiss} provides metric accuracy and geometric precision, but lacks the semantics from RGB input. Our work leverages LIV-SLAM~\cite{zheng2024fast,lin2022r,liu2024omnicolor} to enable efficient, handheld capture of large-scale environments while overcoming the individual limitations of each modality.

\noindent\textbf{Photorealistic sensor simulation}. Traditional approaches rely on textured mesh for sensor simulation ~\cite{chang2017matterport3d,dai2017scannet,xia2018gibson,ramakrishnan2021habitat,dehghan2021arkitscenes,yeshwanth2023scannet++}. While effective for indoor assets, meshes miss fine structures, require aggressive decimation to scale, and cannot encode view-dependent appearance, which limits photorealistic rendering in unbounded scenes. Neural representations like NeRF~\cite{adamkiewicz2022vision,byravan2022nerf2real} and 3DGS~\cite{chen2025splat,lei2025gaussnav,xie2025vid2sim} enable high-quality rendering. However, image-only training introduces scale ambiguity and unstable geometry in neural representations. Surfaces extracted from densities or opacities~\cite{yu2024gaussian,guedon2024sugar} are often noisy or incomplete, weakening collision reliability. Our pipeline combines both worlds: LiDAR provides metric-scale geometry for collision meshes and 3DGS initialization, while 3DGS trained with multi-view images with LIV-SLAM poses enable high-fidelity rendering. This hybrid representation ensures geometrically grounded interaction with photorealistic sensor simulation.

\noindent\textbf{Embodied visual navigation}. A significant body of work in navigation focuses on goal-conditioned navigation~\cite{batra2020objectnav,kadian2020sim2real,krantz2023navigating}. A variety of methods have been developed, ranging from end-to-end Reinforcement Learning~\cite{wijmans2019ddppo} to more structured, modular approaches built upon explicit spatial memories~\cite{savinov2018sptm,chaplot2020neural,zhang2024msg}. Real-world embodied navigation extends these ideas to deploy on quadrupeds or humanoids~\cite{sorokin2022learning,lee2024learning,liu2025citywalker}, but evaluation is often limited to fixed routes or logs due to collection cost, safety, and a lack of open-world, simulator-ready assets. Vision-language navigation (VLN) frames the task as following natural-language instructions. Datasets built from crowd-sourced descriptions~\cite{anderson2018r2r,ku2020rxr,krantz2020vlnce} have advanced this line, yet instruction quality, alignment to visual evidence, and indoor-only layouts are still main constraints. VLN policies range from early sequence models~\cite{krantz2021waypoint} to latest VLA models~\cite{cheng2024navila}. Our dataset supports both families. We generate optimal expert trajectories by computing the shortest path on the navigation mesh. We then leverage MLLM to automatically annotate these traversal videos with rich, contextual descriptions.
\section{\wanderland}
\label{sec:wanderland}

\subsection{Data Collection}

\begin{figure}[t]
    \centering
    \begin{subfigure}[b]{0.37\linewidth}
        \includegraphics[width=\textwidth]{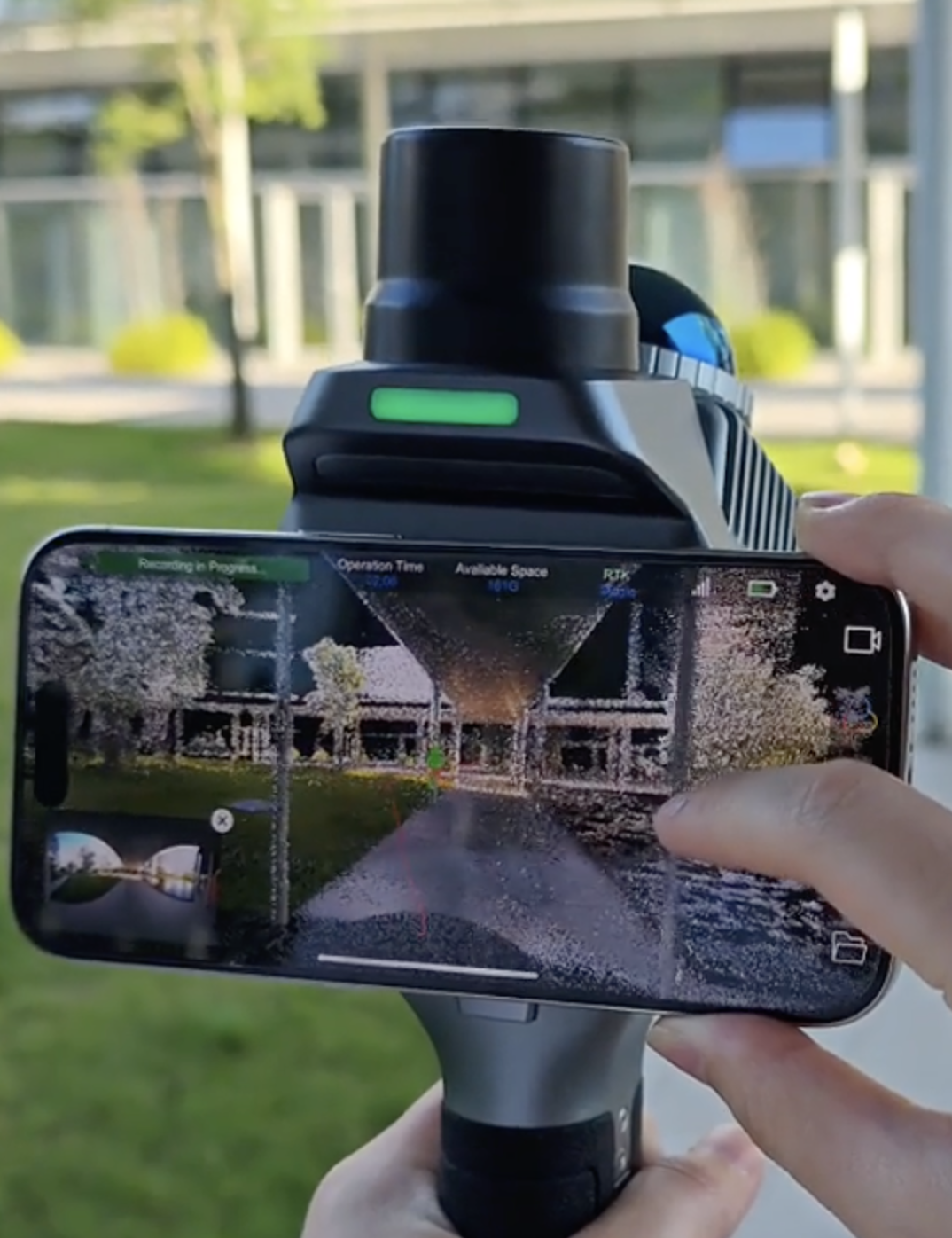}
        \caption{Mobile app.}
    \end{subfigure}
    \hfill
    \begin{subfigure}[b]{0.6\linewidth}
        \includegraphics[width=\textwidth]{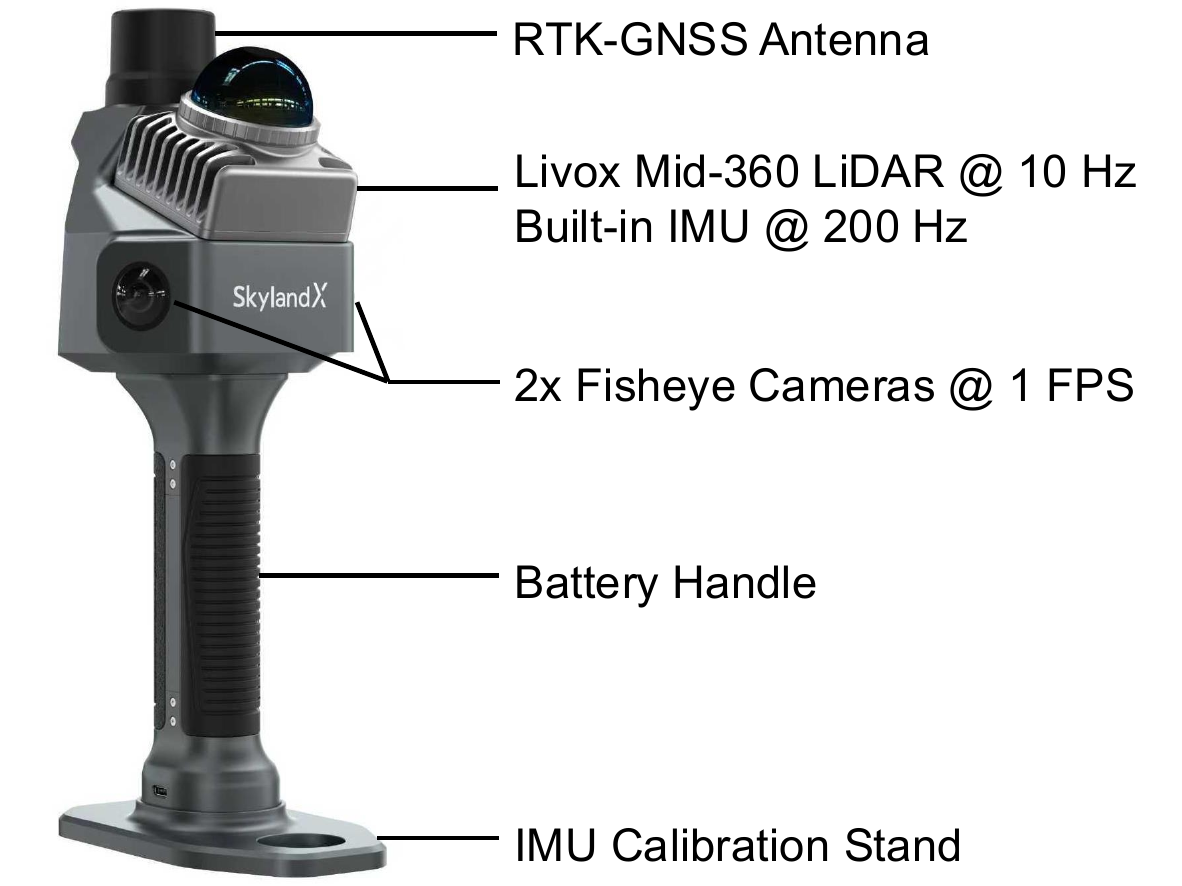}
        \caption{Device specifications.}\label{fig:sub-device}
    \end{subfigure}
    \vspace{-2mm}
    \caption{\textbf{The MetaCam device}. (a) SkylandX MetaCam Air used for data collection, equipped with a companion app for mobile capture. (b) Working frequency of each sensor.}
    \label{fig:metacam}
    \vspace{-4mm}
\end{figure}

\textbf{The MetaCam device}. We collect data using MetaCam\footnote{\url{https://skylandx.com/metacam-air/}}, a compact commercial handheld 3D scanner as illustrated in \cref{fig:metacam}. The device integrates a suite of factory-calibrated sensors: a Livox Mid-360 non-repetitive LiDAR (with a built-in IMU), an RTK-GNSS antenna, and two synchronized 4K fisheye cameras with over 180° field of view. The LiDAR is mounted at a tuned inclination to optimize the capture of ground-level details and maximize the field-of-view (FOV) overlap with the cameras. We also utilize the accompanying mobile app that provides a real-time preview of the colorized point cloud on a mobile device, enabling operators to actively fill any gaps in the environment during capture, ensuring complete scene coverage.

\begin{figure}[t]
    \centering
    \includegraphics[width=0.9\linewidth]{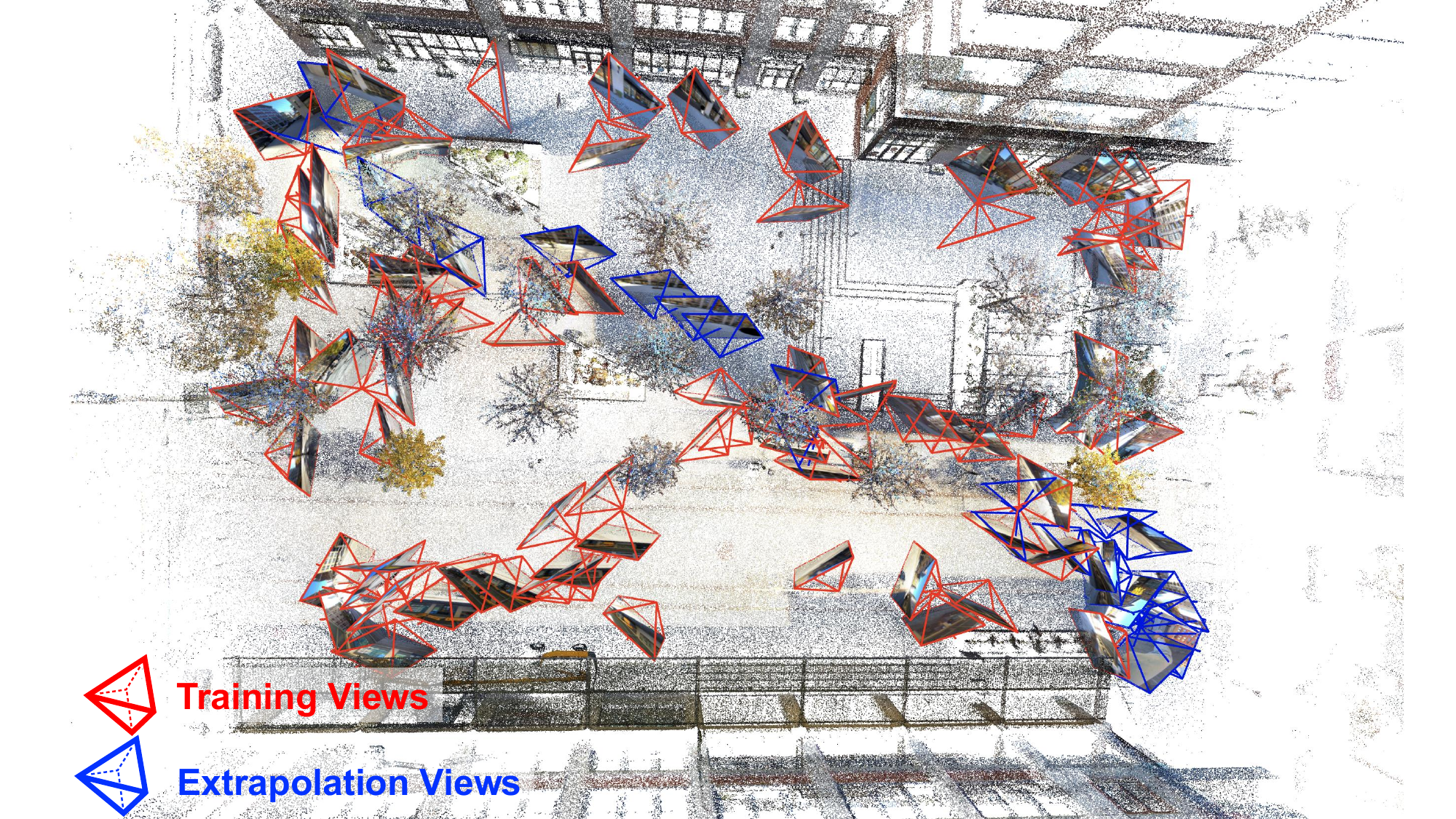}
    \caption{\textbf{Data collection trajectory}. To facilitate both diverse-view capture and evaluation of extrapolated views in navigation, our data is collected with well-defined training and extrapolation splits. Training views ensure accurate reconstruction, while extrapolation views are used for evaluation.}
    \label{fig:capture}
    \vspace{-4mm}
\end{figure}

\noindent\textbf{Collection protocol}. Our data collection scenes are selected to have optimal sizes of 5,000--10,000 square meters to balance complexity and coverage. Instead of recording at a fixed frame rate, we trigger RGB capture whenever the device has moved by a fixed distance or rotated beyond a fixed angular threshold, which leads to more uniformly distributed viewpoints throughout each scene. As shown in \cref{fig:capture}, each capture session employs a systematic trajectory strategy: training trajectories follow closed-loop paths with dense, multi-view coverage of all navigable areas, while extrapolation trajectories simulate natural navigation paths with minimal overlap.  This differs from city touring videos, such as those used in Vid2Sim~\cite{xie2025vid2sim} and GaussGym~\cite{escontrela2025gaussgym}, where the camera views are mostly uni-directional as shown in \cref{fig:teaser} (first column, top two rows). To ensure data quality, we minimize dynamic obstacles and reflective surfaces, maintain consistent lighting conditions during each session, and perform real-time point cloud monitoring to verify coverage completeness.

\noindent\textbf{Time and location}. We collected data across diverse indoor and outdoor urban environments in New York City and Jersey City. Our dataset encompasses distinct scenes spanning residential buildings, business districts, public streets, plazas, and university campuses. More details are shown in Appendix ~\cref{fig:map}. To ensure appearance diversity and robustness, data was captured across different times of day (\textit{e.g.,} morning, noon, dawn) and under varying weather conditions (\textit{e.g.,} sunny, overcast, light rain). 

\begin{figure*}[t]
    \centering
    \includegraphics[width=0.85\linewidth]{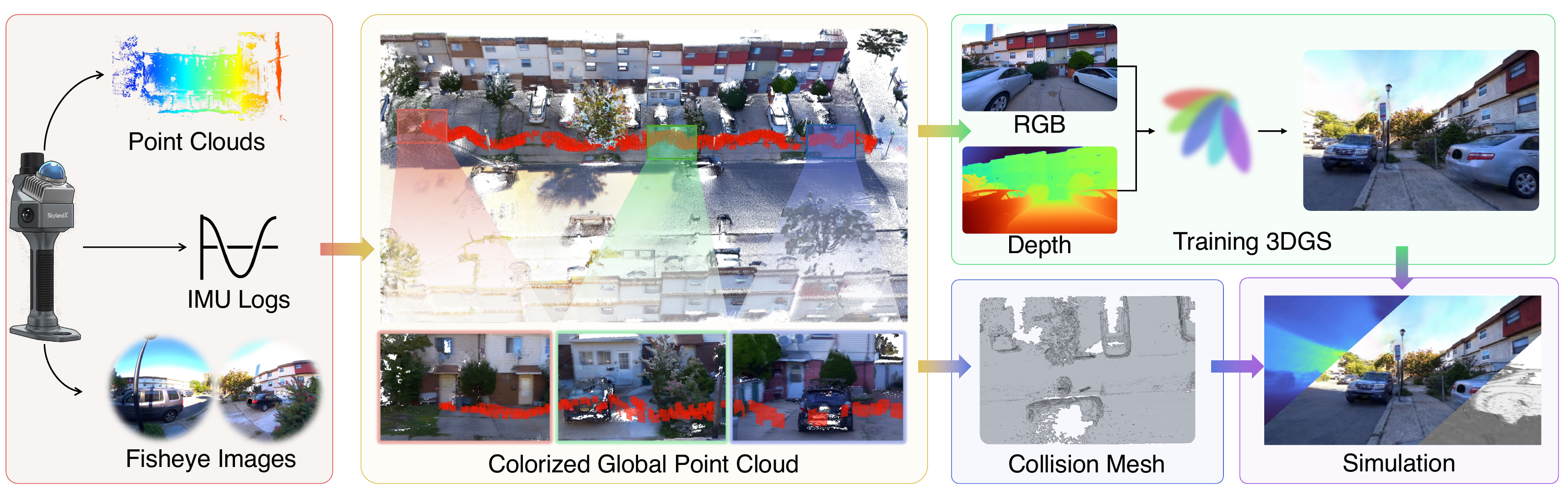}
    \vspace{-2 mm}
    \caption{\textbf{Data Processing Pipeline}. Our pipeline begins with multi-sensor capture using the MetaCam device in real-world urban spaces. MetaCam Studio processes the raw data via LIV-SLAM to produce a colorized, globally consistent metric point cloud and accurate camera poses. We then initialize 3D Gaussians from the metric point cloud and render per-view depth maps from this initialization. The 3DGS model is optimized with both photometric and depth losses. In parallel, we extract a reliable collision mesh from the same global point cloud. Finally, we integrate the trained 3DGS model and the collision mesh into a single Universal Scene Description (USD) scene, which can be directly loaded into Isaac Sim for training and evaluating navigation policies}
    \label{fig:pipeline}
    \vspace{-4mm}
\end{figure*}

\subsection{Data Processing}

\noindent\textbf{Mapping and Reconstruction}. We process the raw sensor data using MetaCam Studio, which implements a robust LiDAR-inertial-visual-GNSS sensor fusion pipeline. While the implementation is proprietary, the methodology builds upon established multi-sensor fusion systems~\cite{qin2018vins,cao2022gvins,zheng2024fast}, extending them for large-scale urban mapping. This sophisticated fusion framework produces dense, metric-scale point clouds with high accuracy and globally consistent camera trajectories as shown in \cref{fig:pipeline}. This provides the geometric foundation for all subsequent steps.

\noindent\textbf{Image masking}. We apply a two-stage masking strategy for our raw fisheye images. First, we use Egoblur~\cite{raina2023egoblur} to mask out human faces and car plates to ensure privacy and anonymity for data release. Second, we use an object detector~\cite{khanam2024yolov11} to mask out common dynamic objects including people, animals, and vehicles, which are used to filter out invalid pixels during 3DGS training.

\noindent\textbf{Image undistortion}. While a large FOV can provide rich coverage of the environment, we crop raw fisheye images into 120\textdegree{} and undistort into perspective views for 3DGS training.  This is due to the limitation from the low-order approximation of camera distortion, and is a common practice in related work~\cite{liao2024fisheye,deng2025self}. It also ensures better masking results as the models are mostly trained on pinhole images. More details in Appendix \cref{sec:supp-3dgs}.

\subsection{Training 3D Gaussians} 

\textbf{Initialization}. We initialize 3D Gaussians from the dense colorized global point cloud produced by MetaCam Studio. Due to its millimeter-level density and accuracy, it provides an high-quality initialization for subsequent 3DGS training. We parameterize Gaussian opacity as inversely proportional to volumetric density obtained by a KNN heuristic. This reduces the dominance of large Gaussians or spurious floaters in initial training steps.

\noindent\textbf{Depth regularization}. Many works have shown depth loss helps 3DGS training~\cite{guedon2024sugar,xie2025vid2sim}. In contrast to use monocular depth as pseudo ground truth, we directly project the initialized Gaussians to each camera pose and treat them as ground truth depth.
While it is intuitive to freeze the Gaussian center attributes considering the accurate initialization, this rigid parameterization imposes an overly uniform geometric prior that limits the model's ability to capture high-frequency details from images, leading to degraded visual quality. We still follow the common practice of combining photometric and depth losses. As a result, we can stabilize 3DGS training and well align the trained Gaussians to the scene geometry.

\noindent\textbf{Training view augmentation}. We also utilize generative models to enhance extrapolated view synthesis. We use the pretrained model from Difix3D+~\cite{wu2025difix3d+} to augment training views with clean and geometrically accurate novel views. We gradually expand the augmented views away from the training views along the training steps. This is crucial for stabilizing large-scale 3DGS training and improving sensor simulation for extrapolated camera views. More details in Appendix \cref{sec:supp-3dgs}. 

\subsection{Geometrically Grounded Simulation}
\textbf{Mesh extraction}. To achieve geometric grounding, we extract mesh from the dense global point cloud. We voxelize the point cloud as an occupancy grid and use the marching cubes~\cite{loresen1987marching} algorithm to obtain geometric meshes. To further improve rendering performance, we remove parts that far from the collection trajectory and filter out fragments with small number of faces. 

\noindent\textbf{Scene integration}. As both the mesh and 3DGS model are based on the global point cloud and share the same coordinate system, we can integrate them into the Unified Scene Description (USD) format. In this representation, the mesh provides a lightweight physics and collision layer, while the 3DGS model is used as the primary renderer. The scene can be directly loaded into Isaac Sim for training and evaluating embodied navigation systems.

\begin{table*}[t]
    \caption{\textbf{Vision-based reconstruction still underperforms LIV-SLAM}. All methods are evaluated on the same number of images for each scene. T-ATE\textsuperscript{R} (raw) and T-ATE\textsuperscript{S} (scaled) means ATE evaluated \textit{without/with} ground-truth scale alignment. COLMAP\textsuperscript{calib} means COLMAP with ground truth intrinsic calibration. SR stands for success rate. Detailed definition on evaluation metrics is described in Appendix \cref{sec:supp-3d}. For each entry, we show the \textbf{mean/median} across the test dataset.}
    \label{tab:exp-3d}
    \vspace{-2mm}
    \centering
    \resizebox{0.9\linewidth}{!}{
    \begin{tabular}{lccccccccc}
        \toprule
        \textbf{Method} & \textbf{T-ATE\textsuperscript{R} (m)} $\downarrow$ & \textbf{T-ATE\textsuperscript{S} (m)} $\downarrow$ & \textbf{R-ATE ($^\circ$)} $\downarrow$ & \textbf{T-RTE (m)} $\downarrow$ & \textbf{T-RTE ($^\circ$)} $\downarrow$ & \textbf{R-RTE ($^\circ$)} $\downarrow$ & \textbf{AUC@30} $\uparrow$ & \textbf{SR} $\uparrow$ \\
        \midrule
        DUSt3R~\cite{wang2024dust3r} &  15 / 14 & 20 / 18 & 73 / 60 & 21 / 20 & 75 / 76 & 75 / 79 & 0.12 / 0.07 & 0.39 \\
        MUSt3R~\cite{cabon2025must3r} &  7.8 / 5.7 &  10 / 3.7 &  26 / 13 &  11 / 8.0 &  37 / 27 &  31 / 17 &  0.53 / 0.61 &  0.81 \\
        VGGT~\cite{wang2025vggt} & 15 / 14 & 9.9 / 4.5 & 33 / 15 & 22 / 20 & 43 / 32 & 35 / 21 & 0.44 / 0.52 & 0.80 \\
        $\pi^3$~\cite{wang2025pi} & 15 / 14 & 4.7 / 1.4 & 21 / 6.9 & 21 / 20 & 26 / 17 & 24 / 8.5 & 0.64 / 0.76 & 0.89 \\
        MapAnything~\cite{keetha2025mapanything} &  6.1 / 4.2 & 8.3 / 4.1 & 30 / 13 &  8.6 / 6.0 & 37 / 30 & 34 / 18 & 0.50 / 0.59  &  0.88 \\
        DA3~\cite{lin2025depth} & 4.9 / 2.8 & 6.0 / 2.3 & 28 / 15 & 6.9 / 3.9 & 32 / 23 & 33 / 20 & 0.50 / 0.56 & 0.86 \\
        COLMAP~\cite{schonberger2016structure} & 16 / 10 &  8.1 / 2.3 &  42 / 10 & 23 / 14 &  38 / 25 &  28 / 12 &  0.50 / 0.64 & 0.64 \\
        COLMAP\textsuperscript{calib}~\cite{schonberger2016structure} &  10 / 9.7 &  4.8 / 0.30 &  15 / 5.0 &  15 / 13 &  16 / 7.7 &  15 / 5.4 &  0.73 / 0.83 &   0.87 \\
        \midrule
        \textbf{Best of All} & 2.8 & 0.30 & 5.0 & 3.9 & 7.7 & 5.4 & 0.83 & 0.89 \\
        \bottomrule
    \end{tabular}
    }
    \vspace{-4mm}
\end{table*}

\subsection{Defining Navigation Tasks}
\textbf{Expert trajectory}. We derive navigation trajectories from mesh geometry. We import the mesh into Unity and utilize the NavMesh baking API to extract a triangulated navigable surface. We then use the pathfinding module to generate collision-free expert trajectories based on the navmesh. The starting and goal positions are sampled in the vicinity of capturing cameras. These trajectories can support point-goal and image-goal navigation tasks.

\noindent\textbf{Language instruction}. For VLN, we additionally generate navigation instructions for each trajectory. We replay each trajectory in the simulation and generate an egocentric video. We use a VLM~\cite{comanici2025gemini} to generate natural language instructions based on the video. The instructions are further verified by humans for reliability. Compared to fully manual annotation~\cite{anderson2018r2r,ku2020rxr}, this procedure is substantially more scalable and yields more stable instructions across scenes, while still allowing for quality control. Detailed steps are described in Appendix \cref{sec:supp-vln}.

\subsection{Data Statistics}
At the current stage, the \wanderland{} dataset comprises 530 distinct scenes with over 420,000 frames captured across more than 100 hours of recording, covering a total area of more than 3.8 million square meters. Each scene provides a comprehensive suite of raw and processed data including: synchronized RGB fisheye images with intrinsic calibrations, globally consistent camera poses, colorized metric point clouds, optimized 3D Gaussian Splatting models, extracted collision meshes, and ready-to-simulate USD scenes. To support long-term research growth, we are committed to continuous dataset maintenance and expansion, with an active development roadmap targeting over 1,000 scenes to further enhance diversity and scale for the embodied AI and 3D computer vision community.

\section{Experiments}
\label{sec:experiments}

We conduct extensive experiments to answer several critical question at the core of this work: 

\noindent \textbf{Q1:} Is the best vision-only 3D reconstruction method as good as LIV-SLAM?

\noindent \textbf{Q2:} Does geometric grounding lead to a better photorealistic simulation quality? 

\noindent \textbf{Q3:} Does video-3DGS framework provide reliable environment for training and evaluation?

\begin{redbox}
\textbf{A1:} \textit{Despite recent bursts in vision-only 3D reconstruction, they have large gaps compared to LIV-SLAM and are not reliable in geometric correctness.}
\end{redbox}

\subsection{3D Reconstruction}
\textbf{Evaluation metrics}. For camera pose estimation, we use standard absolute trajectory error (ATE) and relative trajectory error (RTE) to evaluate the translational (T-) and rotational (R-) accuracy of camera poses. Area under curve at 30\textdegree (AUC@30) and success rate (SR) evaluates overall camera pose accuracy. See \cref{sec:supp-3d} for detailed definition.

\noindent\textbf{Vision only method produces inaccurate camera pose}. In \cref{tab:exp-3d}, our evaluation of camera pose estimation against LIV-SLAM ground truth reveals fundamental limitations in vision only methods. In scenes span less than 100 meters, even if we naively take the ``best of all'' performance, the camera pose estimation accuracy only reaches meter-level metrically. After scale alignment, it still has an average error of 30 cm and 5 degrees. This is due to the inherent modality limitation and the gap is not yet closed by recent foundation models. It prevents the usage of casual videos as data source for reliable reconstruction.

\begin{figure}[t]
    \centering
    \includegraphics[width=0.85\linewidth]{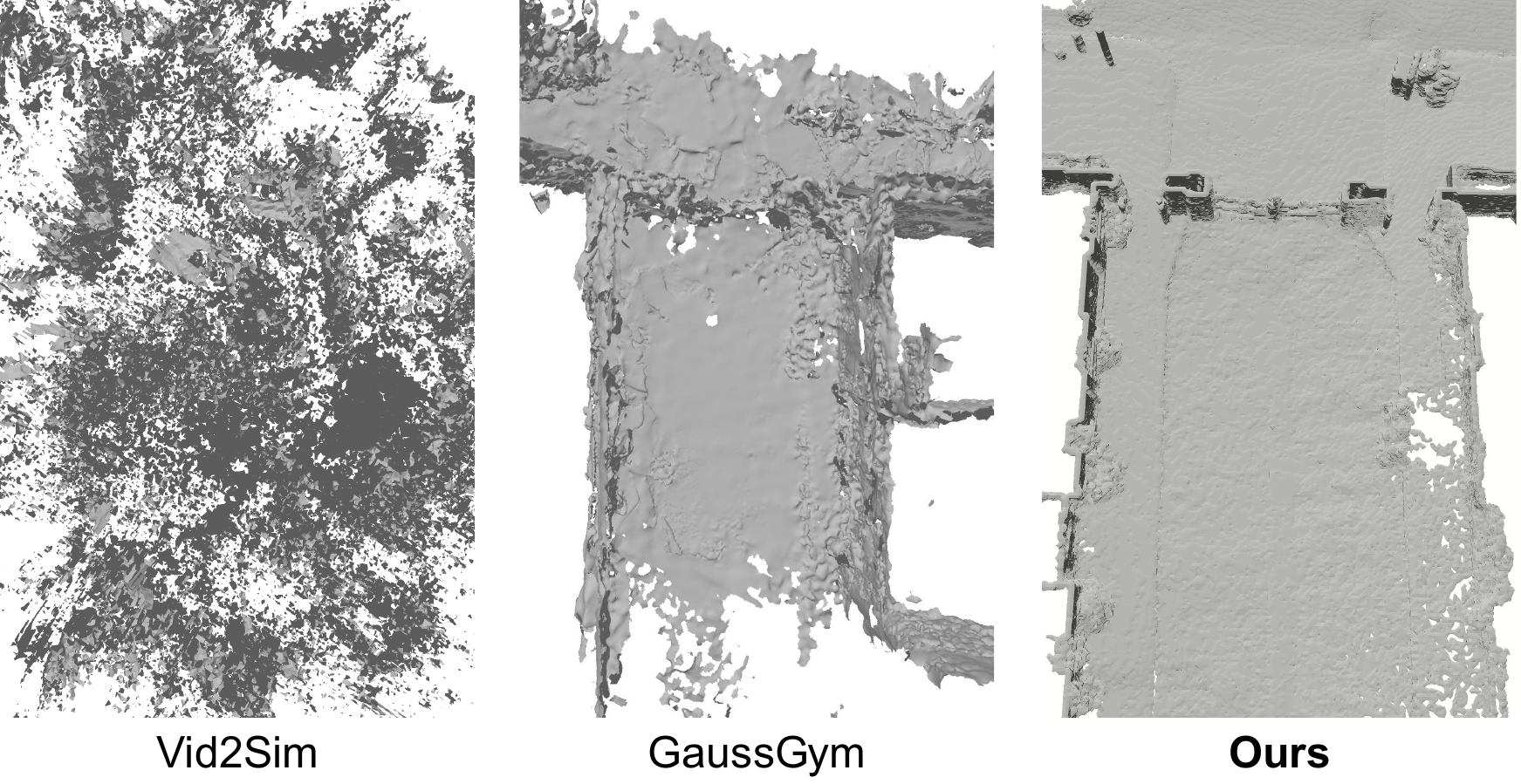}
    \vspace{-3mm}
    \caption{\textbf{Mesh qualitative comparison}. All results are reconstructed from the same data in the \wanderland{} dataset. Our framework extracts complete and smooth mesh.}
    \label{fig:mesh}
    \vspace{-6mm}
\end{figure}

\begin{figure*}[t]
    \centering
    \includegraphics[width=0.85\linewidth]{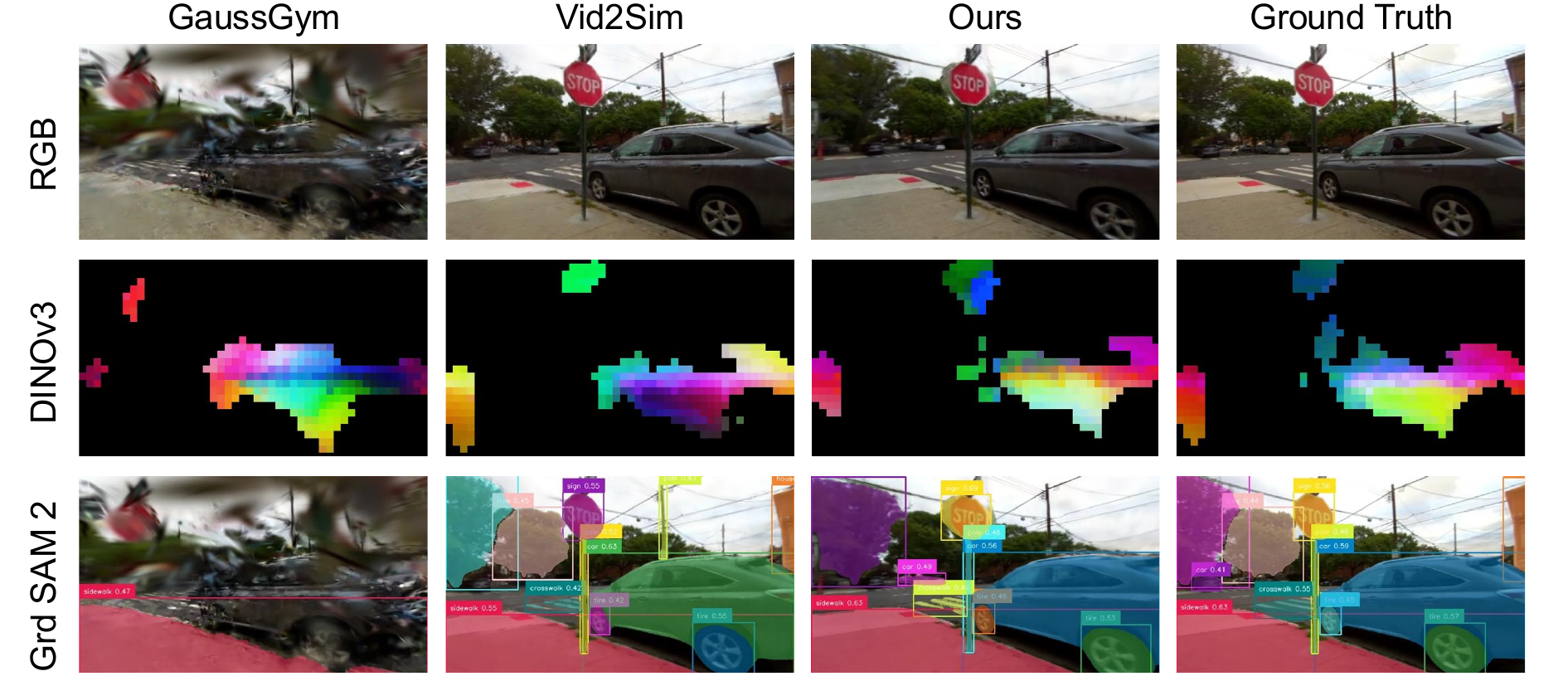}
    \vspace{-4mm}
    \caption{\textbf{Photorealism and semantic consistency for sensor simulation}. We show extrapolated view synthesis results from different frameworks, and their inference results from the DINOv3~\cite{simeoni2025dinov3} and Grounded SAM 2 models~\cite{ren2024grounded}. The DINOv3 visualization shows the first three PCA components on each patch feature. The foreground is filtered by a small linear classifier on DINOv3 features.}
    \label{fig:nvs}
    \vspace{-4mm}
\end{figure*}

\noindent\textbf{Vision only input prevents accurate mesh extraction}. As illustrated in \cref{fig:mesh}, meshes generated from Vid2Sim~\cite{xie2025vid2sim} and GaussGym~\cite{escontrela2025gaussgym} exhibit significant noise, fragmentation, and incompletion, stemming from inaccuracies in the underlying neural representations. These deficiencies undermine collision reliability and hinder physical interaction. In contrast, our method directly extracts meshes from the globally consistent LiDAR point cloud, yielding clean, grounded geometry that ensures metric accuracy and completeness for robust simulation. This comparison underscores the necessity of direct geometric sensing for building actionable environments.

\begin{greenbox}
\textbf{A2:} \textit{Photorealistic sensor simulation significantly benefits from geometric grounding.}
\end{greenbox}

\subsection{Photorealistic Sensor Simulation}
\textbf{Our method outperforms video based frameworks in novel view synthesis}. As shown in ~\cref{tab:exp-nvs}, all baseline methods achieve subpar results. GaussGym~\cite{escontrela2025gaussgym} exhibits the lowest metrics due to the inaccurate 3D reconstruction from VGGT~\cite{wang2025vggt}. Vid2Sim's~\cite{xie2025vid2sim} unsatisfactory performance stems from its reliance on inaccurate monocular depth estimation~\cite{yang2024depth} as supervision, introducing additional noise during training. As for other baselines, their limited performance suggests that photometric losses alone fail to leverage geometric scene information effectively. In contrast, our method achieves superior performance by leveraging LIV-SLAM's accurate poses and geometric initialization, demonstrating that robust NVS requires both visual and geometric foundations.

\begin{table}
    \caption{\textbf{Better NVS from geometric grounding}. All methods are evaluated on our \wanderland{} dataset with the same train and validation split (including both interpolated and extrapolated views.) Vid2Sim~\cite{xie2025vid2sim} uses reconstruction from GLOMAP~\cite{pan2024global} and GaussGym~\cite{escontrela2025gaussgym} uses reconstruction from VGGT~\cite{wang2025vggt}.}
    \label{tab:exp-nvs}
    \vspace{-2mm}
    \centering
    \resizebox{\linewidth}{!}{
    \begin{tabular}{lcccccc}
        \toprule
        \multirow{2}{4em}{\textbf{Method}} & \multicolumn{3}{c}{\textbf{Interpolated Views} } & \multicolumn{3}{c}{\textbf{Extrapolated Views}} \\
        \cmidrule(lr){2-4} \cmidrule(lr){5-7}
         & \textbf{PSNR} $\uparrow$ & \textbf{SSIM} $\uparrow$ & \textbf{LPIPS} $\downarrow$ & \textbf{PSNR} $\uparrow$ & \textbf{SSIM} $\uparrow$ & \textbf{LPIPS} $\downarrow$ \\
        \midrule
        \multicolumn{7}{l}{\cellcolor[HTML]{E8E8E8} \textit{COLMAP Reconstruction}} \\
        3DGS~\cite{kerbl20233d} & \second 18.27 & \second 0.658 & \third 0.510 & \third 16.90 & \second 0.624 & 0.559\\
        2DGS~\cite{huang20242d} & 17.98 & 0.593 & 0.550 & 16.81 & \first 0.631 & \third 0.508 \\
        3DGUT~\cite{wu20253dgut} & \third 18.29 & \third 0.654 & 0.535 & \second 17.00 & \third 0.619 & 0.576 \\
        \midrule
        \multicolumn{7}{l}{\cellcolor[HTML]{E8E8E8} \textit{Custom Reconstruction}} \\
        Vid2Sim~\cite{xie2025vid2sim} & 17.20 & 0.549 & \second 0.399 & 16.49 & 0.573 & \first 0.371 \\
        GaussGym~\cite{escontrela2025gaussgym} & 12.17 & 0.440 & 0.738 & 12.63 & 0.436 & 0.725 \\
        \midrule
        \multicolumn{7}{l}{\cellcolor[HTML]{E8E8E8} \textit{LIV-SLAM Reconstruction}} \\
        \textbf{Ours} & \first 20.37  & \first 0.688 & \first 0.327  & \first 17.92 & 0.591 & \second 0.445 \\
        \bottomrule
    \end{tabular}
    }
    \vspace{-6mm}
\end{table}

\noindent\textbf{Our framework supports semantically consistent sensor simulation}. Beyond standard NVS metrics, we investigate semantic consistency for sensor simulation. As shown in \cref{fig:nvs}, GaussGym's fragmented renderings prevent reliable segmentation, with Grounded SAM 2 failing to detect critical environmental elements. While Vid2Sim produces more coherent images that support detection and segmentation, its DINOv3 features diverge significantly from ground truth (drastically different PCA colors). This can confuse end-to-end navigation policies that rely on DINO features for semantic understanding~\cite{liu2025citywalker}. In contrast, our renderings maintain both structural integrity and semantic consistency, enabling accurate segmentation and producing DINOv3 features closely aligned with real imagery. These results demonstrate that rendering quality directly impacts perception models, and that geometric accuracy is essential for reliable embodied AI in open-world environments.

\begin{bluebox}
\textbf{A3:} \textit{Our framework builds more reliable environment for both training and evaluating embodied navigation.}
\end{bluebox}

\begin{table}
    \caption{\textbf{Geometrically grounded environment for RL training}. We compare model performance and their change after post-trained on different simulation environments. Unlike \cref{tab:exp-nav}, the results are evaluated on the test split of the \wanderland{} dataset. Numbers inside parentheses indicate metric \textbf{change} compared to pretrained model. We use background color to represent model \colorbox{positivebg}{\textbf{improvement}} and \colorbox{negativebg}{\textbf{deterioration}} after the RL training.}
    \label{tab:exp-rl}
    \vspace{-2mm}
    \centering
    \resizebox{\linewidth}{!}{
    \begin{tabular}{llcccc}
        \toprule
        \textbf{RL Env.} & \textbf{Methods} & \textbf{NE (m)} $\downarrow$ & \textbf{SR} $\uparrow$ & \textbf{SPL} $\uparrow$ & \textbf{IR} $\downarrow$ \\
        \midrule
        \multirow{3}{5em}{Vid2Sim~\cite{xie2025vid2sim}} 
            & NoMaD~\cite{sridhar2024nomad}      
                & \negc 19.11 (+4\%)  & 0.24 (0\%)   & \negc 0.23 (-4\%) & \posc 0.64 (-9\%) \\
            & CityWalker~\cite{liu2025citywalker} 
                & \negc 26.40 (+24\%) & \negc 0.17 (-21\%) & \negc 0.17 (-19\%) & \posc 0.64 (-11\%) \\
            & MBRA~\cite{hirose2025learning}     
                & \posc 23.76 (-4\%)  & 0.22 (0\%)   & \negc 0.21 (-5\%) & \negc 0.69 (+1\%) \\
        \midrule
        \multirow{3}{5em}{\textbf{Ours}} 
            & NoMaD~\cite{sridhar2024nomad}      
                & \posc 17.99 (-2\%)  & \posc 0.26 (+8\%) & 0.24 (0\%)   & \posc 0.61 (-13\%) \\
            & CityWalker~\cite{liu2025citywalker} 
                & \posc 19.02 (-10\%) & \posc 0.24 (+14\%) & \posc 0.23 (+14\%) & \posc 0.55 (-23\%) \\
            & MBRA~\cite{hirose2025learning}     
                & \posc 23.76 (-4\%)  & \posc 0.23 (+5\%) & \posc 0.23 (+5\%) & \posc 0.63 (-7\%) \\
        \bottomrule
    \end{tabular}
    }
    \vspace{-6mm}
\end{table}

\subsection{Embodied Navigation}
\label{sec:experiments-nav}

\textbf{Evaluation metrics}. We use standard navigation error (NE), success rate (SR), and success path length (SPL) to evaluate embodied navigation performance. We additionally defined the intervention rate (IR) to evaluate none-timeout failure cases. We describe more details on metric definition and experiment setup in Appendix \cref{sec:supp-navigation}.

\noindent\textbf{Reinforcement learning with geometric grounding}. In \cref{tab:exp-rl}, we perform reinforcement learning (RL) training in environments built from different frameworks. Notably, models generally deteriorate after training in environments built by Vid2Sim~\cite{xie2025vid2sim}. Under inaccurate geometry, the RL objective encourages locally shorter but globally unreliable behaviors, which results in lower success rate during evaluation. In contrast, all models significantly improve when trained in our geometrically grounded environments. This comparison highlights that RL fine-tuning is highly sensitive to the underlying simulation fidelity: metrically accurate, collision-consistent geometry is crucial for RL to produce genuinely better navigation policies.

\begin{figure}[t]
    \centering
    \includegraphics[width=\linewidth]{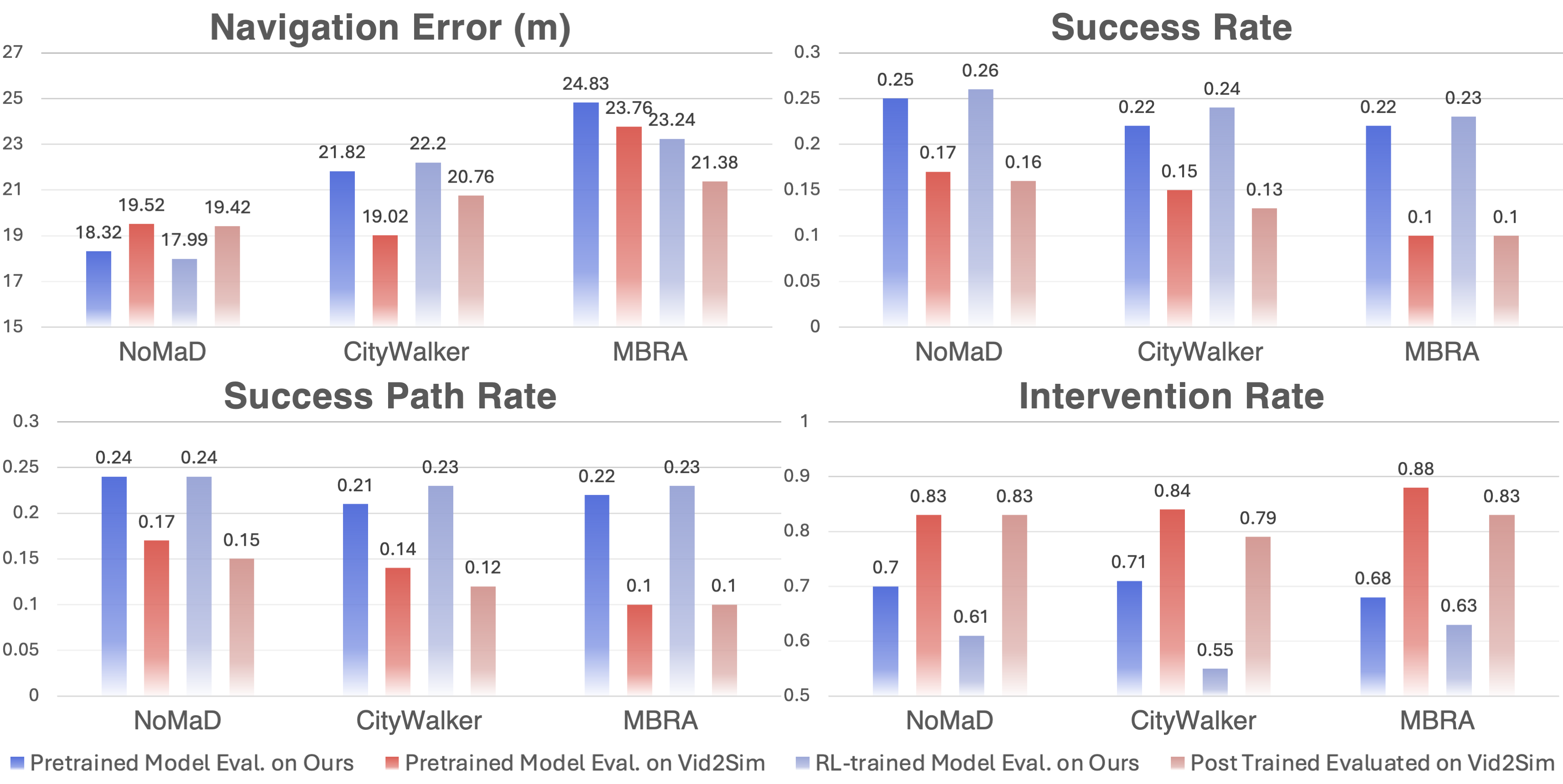}
    \caption{\textbf{Geometrically grounded environment for evaluation}. We compare evaluation results of the same models on different environments. Results are evaluated on the \wanderland{} test split. Comparison should be made between different evaluation environments (different colors) for same models.}
    \label{fig:exp-dataset}
    \vspace{-5mm}
\end{figure}

\noindent\textbf{Evaluation with geometric grounding}. We further demonstrate that model evaluation can be unreliable in the absence of grounded geometry. As shown in \cref{fig:exp-dataset}, when evaluated with Vid2Sim-built environments (red bars), all models show a much lower success rate and higher intervention rate compared to our environments (blue bars). This indicates that geometrically unreliable environments fail to support a faithful evaluation. To minimize the sim-to-real gap, geometrically grounded simulation is a must in benchmarking embodied navigation policies.

\noindent\textbf{Benchmarking pretrained navigation policies}. \Cref{tab:exp-nav} summarized the performance of pretrained models evaluated on the \wanderland{} dataset. Our first observation is that VLN models generally outperforms point-goal and image-goal models. This is expected as VLN is now considered a less challenging task compared to others (only with the recent help of LLMs). Another observation is that outdoor navigation is generally a more challenging task than indoor scenes. This is due to the longer trajectories, more complex topology, and elevation changes. Overall, the benchmark shows a large research gap in open-world embodied navigation as none of the models reach a success rate over 50\%.

\begin{table}
    \caption{\textbf{Navigation Benchmark}. We benchmark different navigation models for different tasks on our entire \wanderland{} dataset. See Appendix \cref{sec:supp-navigation} for detailed metric definition.}
    \label{tab:exp-nav}
    \vspace{-2mm}
    \centering
    \resizebox{\linewidth}{!}{
    \begin{tabular}{lcccccccc}
        \toprule
        \multirow{2}{4em}{\textbf{Methods}} & \multicolumn{4}{c}{\textbf{Indoor Scenes}} & \multicolumn{4}{c}{\textbf{Outdoor Scenes}} \\
        \cmidrule(lr){2-5} \cmidrule(lr){6-9}
         & \textbf{NE (m)} $\downarrow$ & \textbf{SR} $\uparrow$ & \textbf{SPL} $\uparrow$ & \textbf{IR} $\downarrow$ & \textbf{NE (m)} $\downarrow$ & \textbf{SR} $\uparrow$ & \textbf{SPL} $\uparrow$ & \textbf{IR} $\downarrow$ \\
         \midrule
        NoMaD~\cite{sridhar2024nomad} & \third 7.04 & 0.22 & 0.22 & \second 0.52 & \second 13.4 & \second 0.24 & \second 0.24 & \third 0.70 \\
        CityWalker~\cite{liu2025citywalker} & 7.82 & \second 0.39 & \second 0.39 & 0.59 & \third 16.4 & 0.21 & 0.20 & 0.72 \\
        MBRA~\cite{hirose2025learning} & \second 5.28 & \third 0.35 & \third 0.34 & \third 0.55 & 19.4 & \third 0.22 & \third 0.22 & \first 0.68 \\
        NaVid~\cite{zhang2024navid} & 15.1 & 0.29 & 0.25 &  0.66 & 28.5 & 0.15 & 0.14 &  0.77 \\
        NaVILA~\cite{cheng2024navila} & \first 5.13 & \first 0.47 & \first 0.47 & \first 0.41 & \first 13.2 & \first 0.31 & \first 0.31 & \first 0.68 \\
        \bottomrule
    \end{tabular}
    }
    \vspace{-4mm}
\end{table}

\section{Discussion and Conclusion}
\label{sec:discussion}

\textbf{Broader Impact}. Beyond bridging the real-to-sim gap for embodied navigation, \wanderland{} provides foundational resources for multiple research domains. The metric-scale camera poses and dense LiDAR point clouds offer unprecedented ground-truth data for foundational vision geometry models. The carefully designed extrapolated views enable rigorous benchmarking of novel view synthesis methods under realistic off-trajectory conditions. Critically, our dataset addresses the scarcity of large-scale metric benchmarks for outdoor 3D vision, enabling new research directions in long-range depth estimation and geometric learning. By providing reliable geometric ground truth at scale, \wanderland{} establishes a new standard for research in 3D computer vision and embodied AI.

\noindent\textbf{Limitations}. Our framework has two primary limitations. First, the current capture system operates at 1 FPS due to hardware constraints, resulting in sparser viewpoint sampling than ideal. This limits the density of training views and consequently affects the final rendering quality in highly complex scenes. We plan to address this through hardware upgrades in our ongoing dataset development. Second, while we focus on geometric reconstruction and static environment simulation, real urban environments involve complex dynamics including moving pedestrians, vehicles, and traffic patterns. Modeling these dynamic elements remains an important challenge for future work, requiring integration with behavior prediction and interactive simulation beyond the scope of this paper.

\noindent\textbf{Conclusion}. We demonstrate that reliable simulation for open-world embodied AI requires geometrically grounded environments, which is a critical requirement unmet by current video-based 3DGS pipelines. Our work shows that vision-only reconstruction fails to deliver the metric accuracy, reliable geometry, and consistent view synthesis needed for reproducible benchmarking. By introducing the \wanderland{} framework and dataset, we establish a new foundation for embodied AI research, where perception, planning, and evaluation can be built upon accurate and scalable simulations. Moving forward, we argue that geometric grounding is not optional, but essential for the next generation of embodied systems operating in real world.

\section*{Acknowledgment}
The work was supported in part through NSF grants 2514030, 2238968, and 2345139, in part by NVIDIA Academic Grant Program, and the NYU IT High Performance Computing resources, services, and staff expertise. We thank SkylandX for their technical support. We thank Hellon Luo, Shiqi Wang, Ying Wang, Zhicheng Yang, and Yining Zheng for their help in data collection. We thank Juexiao Zhang and Sihang Li for insightful discussion.

{
    \small
    \bibliographystyle{unsrt}
    \bibliography{main}
}
\clearpage
\setcounter{section}{0}
\setcounter{figure}{0}
\setcounter{table}{0}
\renewcommand{\thesection}{\Alph{section}}
\renewcommand{\thefigure}{\Roman{figure}}
\renewcommand{\thetable}{\Roman{table}}

\section*{Appendix}
\section{Data Collection and Processing Details}
\label{sec:supp-collection}
\begin{figure}[t]
    \centering
    \includegraphics[width=\linewidth]{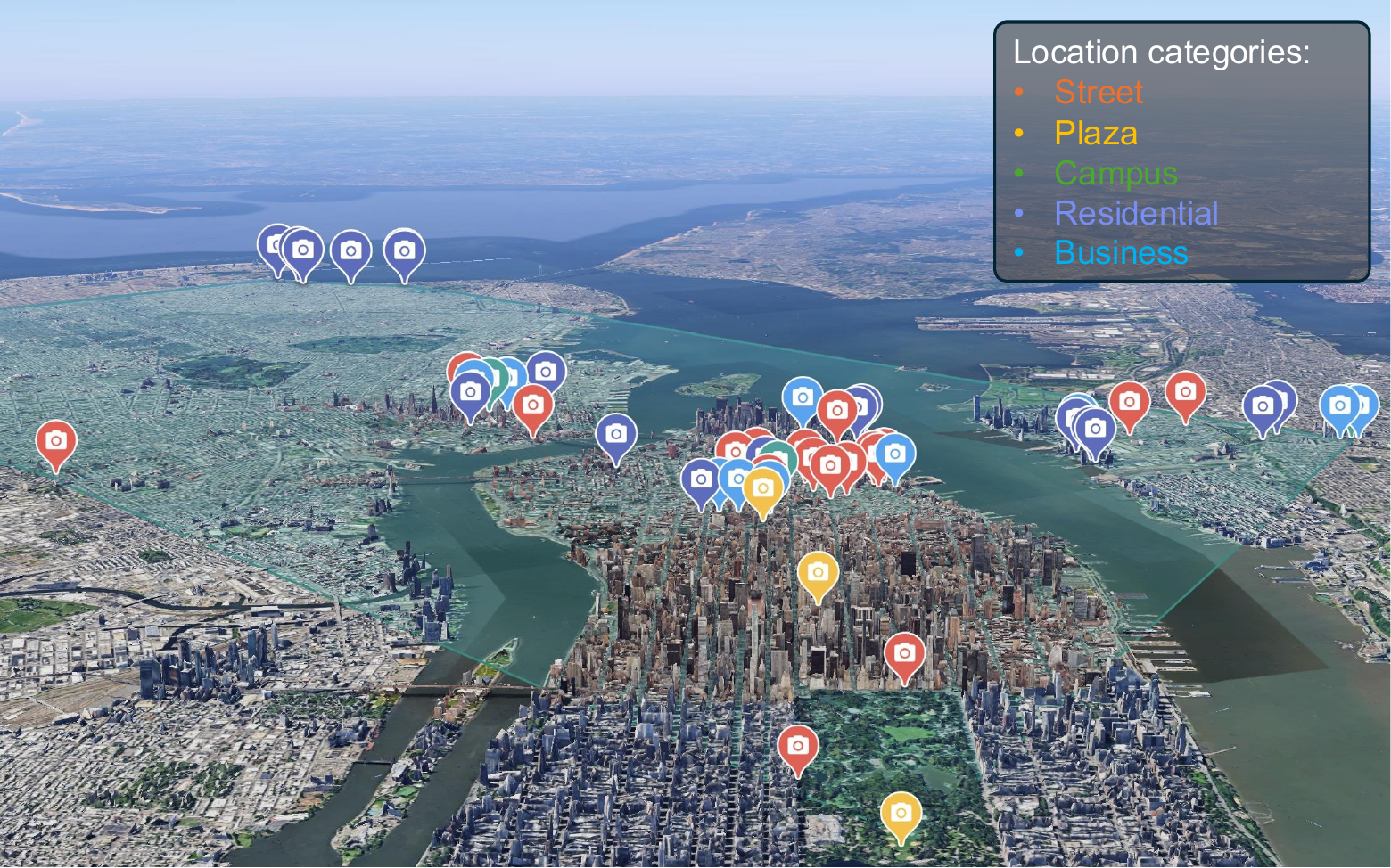}
    \caption{\textbf{Data collection locations}. We collect data in New York City and Jersey City. The capture location is carefully selected to cover different scenes.}
    \label{fig:map}
\end{figure}

\Cref{fig:map} shows a selection of our data collection locations. Different colors incidate different location types.

\begin{figure*}
    \centering
    \includegraphics[width=\linewidth]{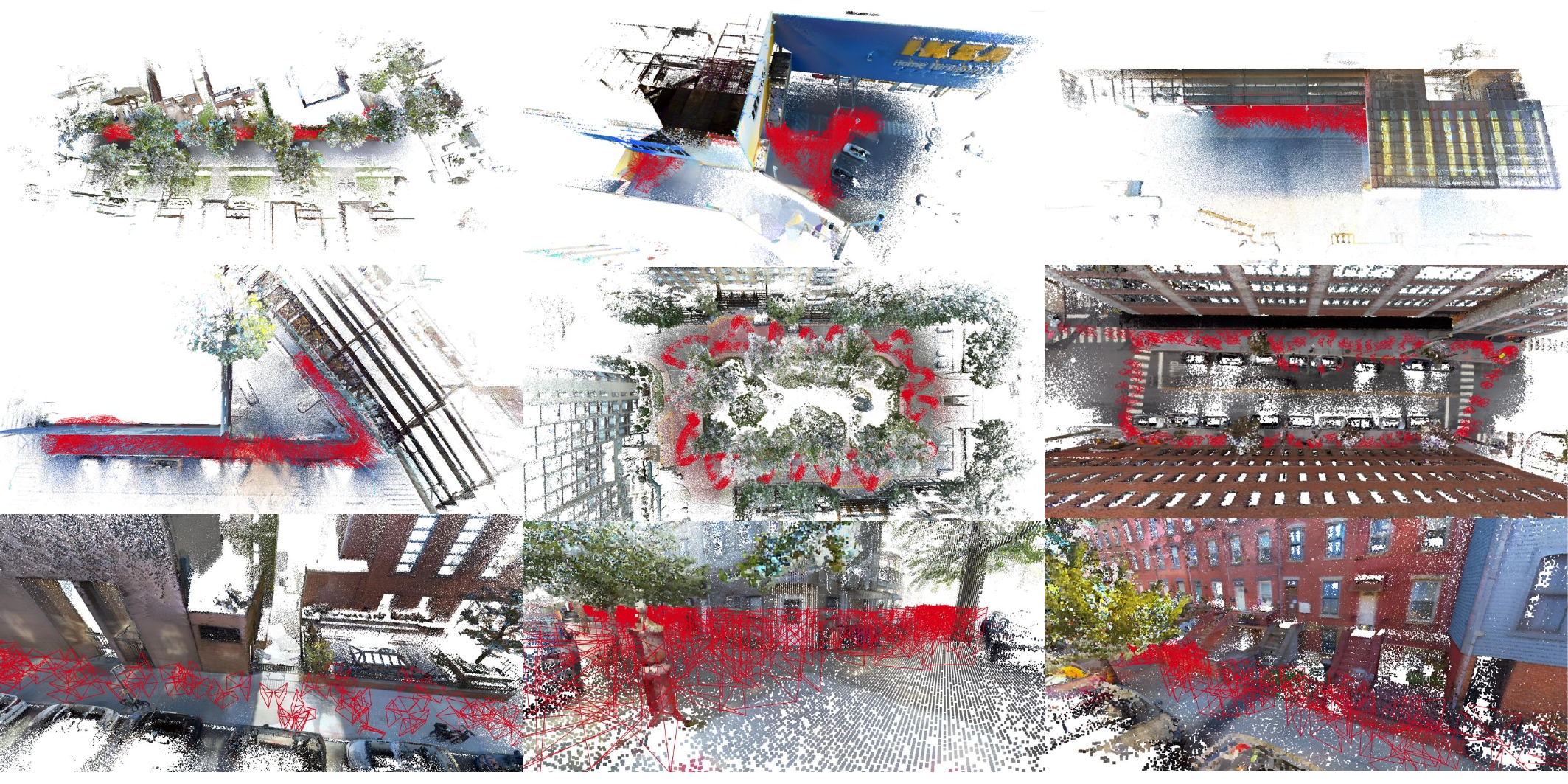}
    \caption{\textbf{Our 3D reconstruction from LIV-SLAM}. The global colorized point cloud is downsampled for visualization cleanness.}
    \label{fig:supp-3drecon}
\end{figure*}

\noindent \Cref{fig:supp-3drecon} shows more visualization on our 3D reconstruction results.

\section{3DGS Training Details}
\label{sec:supp-3dgs}

\textbf{Training Setup}. Our 3DGS implementation is built on top of the open-source \texttt{gsplat}~\cite{ye2025gsplat} framework, which provides an efficient and scalable renderer for Gaussian splatting.  For all experiments, we render and train at a fixed resolution of 800$\times$800 pixels. This resolution offers a good balance between spatial detail and GPU memory usage, and allows us to handle large urban scenes without exhausting GPU memory. Other hyperparameters for 3DGS training is listed in \cref{tab:hyperparameter}.

\noindent \textbf{Initialization from metric point clouds}. 
For each scene, we initialize 3D Gaussians from the dense colorized point cloud produced by MetaCam Studio. The raw point cloud has a spacing of 5--10\,mm, resulting in roughly 10--50 million points per scene. 
To keep training tractable while preserving sufficient detail for a five-minute walking-scale street scene, we uniformly downsample the point cloud to around 5 million points per scene and create one Gaussian per point. Our initialization largely follows the default settings in \texttt{gsplat}: we use a k-nearest-neighbor heuristic to set the initial \emph{scale} of each Gaussian, and parameterize initial opacity inversely proportional to initial volume. 
This density-based parameterization prevents large Gaussians or spurious floaters left by transient obstacles from artificially dominating the rendering at the beginning of training.

\noindent \textbf{Depth Regularization}.
Because the global point cloud is extremely dense and globally consistent, the resulting depth maps rendered from initialized Gaussians provide a close approximation to ground-truth geometry. This term acts as a geometric prior: it regularizes Gaussians along the viewing rays and prevents them from drifting into free space or collapsing towards the cameras during optimization.

A natural alternative, enabled by accurate LiDAR point cloud, is to freeze the Gaussian means and only optimize appearance-related parameters such as color, opacity, and rotation. We experimented with this stronger form of supervision and found that, although it indeed preserves excellent multi-view geometric consistency, it yields suboptimal visual quality and can still produce degenerate behavior around training views, as shown in \cref{tab:supp-geom-compare}. 
In contrast, relying purely on image supervision from a limited set of views improves per-view fidelity but tends to sacrifice consistency for unseen viewpoints. Our depth regularization strikes a balance between these extremes: it keeps the learned geometry close to the dense metric point cloud while still allowing Gaussians to move and adapt to fit high-quality images.

\begin{table}[t]
    \caption{Hyperparameters for our 3DGS training.}
    \label{tab:hyperparameter}
    \centering
    \resizebox{0.7\linewidth}{!}{
    \begin{tabular}{lc}
        \toprule
        \textbf{Hyperparameter} & \textbf{Value} \\
        \midrule
        Camera model & Pinhole \\
        Camera FOV & 120$^\circ$ \\
        Image resolution & 800$\times$800 \\
        Training steps & 15{,}000 \\
        SH degree & 3 \\
        Initial opacity & 0.99 \\
        Initial scale & 0.5 \\
        Perceptual loss weight & 0.2 \\
        Depth loss weight & 0.02 \\
        Means learning rate & $1.6 \times 10^{-5}$ \\
        Scales learning rate & $1.0 \times 10^{-3}$ \\
        Opacity learning rate & $2.0 \times 10^{-2}$ \\
        Quaternion learning rate & $1.0 \times 10^{-3}$ \\
        SH band 0 learning rate & $5.0 \times 10^{-4}$ \\
        SH band N learning rate & $1.25 \times 10^{-4}$ \\
        Opacity regularization & 0 \\
        Scale regularization & 0.01 \\
        \bottomrule
    \end{tabular}
    }
    \vspace{-2mm}
\end{table}

\begin{table}[t]
    \caption{\textbf{Comparison of geometric priors}. 
    We compare our depth-based regularization with the alternative strategy of freezing Gaussian centers to the LiDAR point cloud. 
    }
    \label{tab:supp-geom-compare}
    \centering
    \vspace{-2mm}
    \resizebox{\linewidth}{!}{
    \begin{tabular}{lccc ccc}
        \toprule
        \multirow{2}{*}{\textbf{Geometric Prior}} 
        & \multicolumn{3}{c}{\textbf{Interpolated Views}} 
        & \multicolumn{3}{c}{\textbf{Extrapolated Views}} \\
        \cmidrule(lr){2-4} \cmidrule(lr){5-7}
        & \textbf{PSNR} $\uparrow$ & \textbf{SSIM} $\uparrow$ & \textbf{LPIPS} $\downarrow$ 
        & \textbf{PSNR} $\uparrow$ & \textbf{SSIM} $\uparrow$ & \textbf{LPIPS} $\downarrow$ \\
        \midrule
        Depth Loss
            & 20.37 & 0.688 & 0.327 
            & 17.92 & 0.591 & 0.445 \\
        Frozen Gaussians 
            & 20.39 & 0.703 & 0.327 
            & 17.10 & 0.558 & 0.456 \\
        \bottomrule
    \end{tabular}
    }
    \vspace{-4mm}
\end{table}

\begin{table*}[t]
    \caption{\textbf{Ablation study on 3DGS training components}. ``P'' stands for pinhole and ``F'' stands for fisheye.}
    \label{tab:supp-nvs-ablation}
    \vspace{-2mm}
    \centering
    \resizebox{0.9\linewidth}{!}{
    \begin{tabular}{cccccc ccc ccc}
        \toprule
        \multicolumn{6}{c}{3DGS Components} & \multicolumn{3}{c}{Interpolated Views } & \multicolumn{3}{c}{Extrapolated Views} \\
        \cmidrule(lr){1-6} \cmidrule(lr){7-9} \cmidrule(lr){10-12} 
        \rotatebox{75}{\textbf{Diverse Views}} & \rotatebox{75}{\textbf{GT Camer Pose}} & \rotatebox{75}{\textbf{LiDAR Point Cloud}} & \rotatebox{75}{\textbf{Depth Loss}} & \rotatebox{75}{\textbf{View Augmentation}} &
        \rotatebox{75}{\textbf{Camera Model}} &\textbf{PSNR} $\uparrow$ & \textbf{SSIM} $\uparrow$ & \textbf{LPIPS} $\downarrow$ & \textbf{PSNR} $\uparrow$ & \textbf{SSIM} $\uparrow$ & \textbf{LPIPS} $\downarrow$ \\
        \midrule
         & & & & & P & 15.26 & 0.515 & 0.558 & 12.99 & 0.457 & 0.655 \\
        $\checkmark$ & & & & & P & 18.27 & 0.658 & 0.510 & 16.90 & 0.624 & 0.559 \\
        $\checkmark$ & $\checkmark$ & & & & P & 15.14 & 0.565 & 0.693 & 14.66 & 0.546 & 0.716 \\
        $\checkmark$ & $\checkmark$ & $\checkmark$ & & & P & 15.54 & 0.532 & 0.605 & 15.19 & 0.515 & 0.625 \\
        $\checkmark$ & $\checkmark$ & $\checkmark$ & $\checkmark$ & & F & \second 20.91 & \third 0.681 & \first 0.262 & \third 16.87 & \third 0.530 & \third 0.468\\
        $\checkmark$ & $\checkmark$ & $\checkmark$ & $\checkmark$ & & P & \first 20.95 & \first 0.696 & \second 0.269  & \second 16.97 & \second 0.544 & \second 0.452\\
        $\checkmark$ & $\checkmark$ & $\checkmark$ & $\checkmark$ & $\checkmark$ & P & \third 20.37 & \second 0.688 & \third 0.327 & \first 17.92 & \first 0.591 & \first 0.445\\
        \bottomrule
    \end{tabular}
    }
    \vspace{-2mm}
\end{table*}

\noindent \textbf{Training view augmentation}. We utilize the pretrained Difix3D+~\cite{wu2025difix3d+} model to augment our training camera views. 
Instead of modifying any original dataset images, we follow a self-training strategy inspired by Difix3D+: for a sampled extrapolated camera pose, we first rasterize an image from the current 3DGS model, then feed it together with its nearest neighboring training views as references into Difix3D+ to obtain a cleaned and geometrically accurate novel view. The synthesized image is then added back into the training set and supervised with a lower loss weight, so that it doesn't the original observations.

\begin{figure*}
    \centering
    \includegraphics[width=\linewidth]{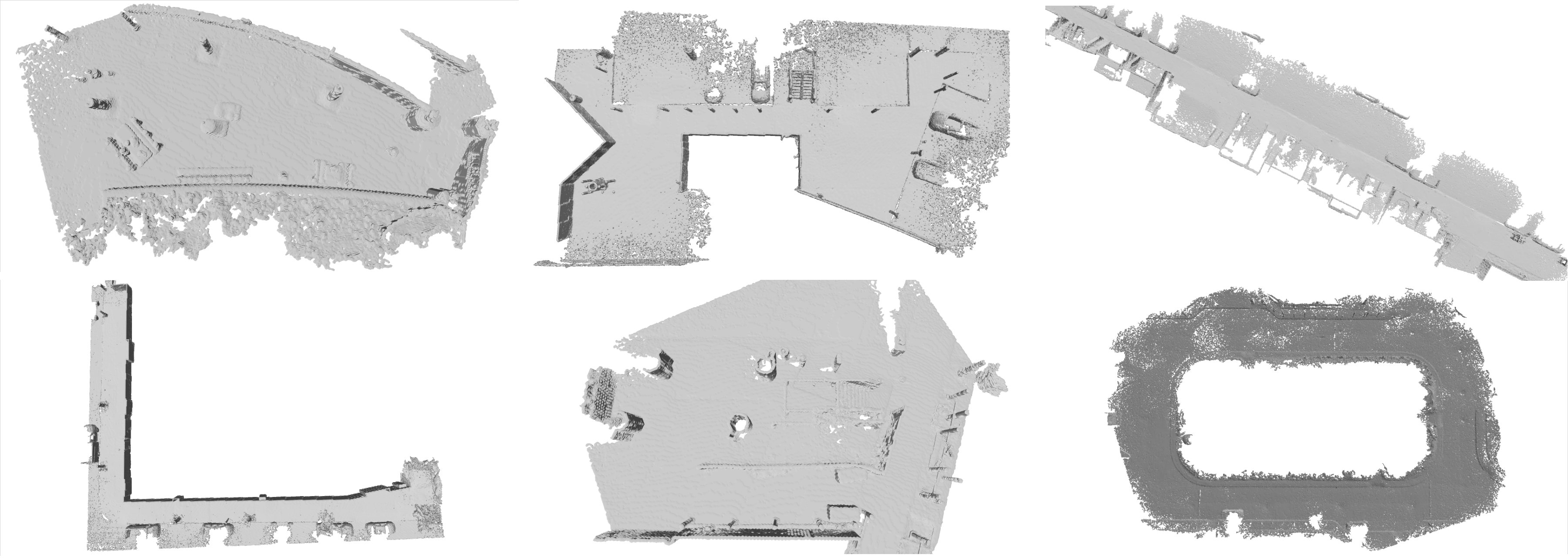}
    \caption{\textbf{Our mesh extraction from global point cloud}. The mesh is cut by a height threshold and a radius threshold around capturing cameras. This ensures a lightweight mesh while keeping geometry consistency.}
    \label{fig:supp-mesh}
    \vspace{-4mm}
\end{figure*}

Our augmentation strategy is different from that in Difix3D+, which mainly densifies views along specific paths to improve rendering quality near interpolated evaluation trajectories. We adopt a more general sampling strategy tailored to large-scale navigation scenes. At the early stages of training, we only sample and refine novel views in a small neighborhood around the training trajectories. Along the training steps, we gradually increase the sampling radius and move the extrapolated viewpoints farther away, effectively implementing a curriculum over viewpoint distance. This procedure increases the diversity and coverage of training views in a controlled manner, which is crucial for stabilizing large-scale 3DGS training and improving generalization to truly unseen viewpoints.

\noindent \textbf{Ablation study}. \Cref{tab:supp-nvs-ablation} studies different 3DGS training design choices. We observe that while ground truth camera pose and dense LiDAR points provides a good initialization, they have to be well utilized by a good regularization signal (\textit{e.g.} the depth loss) to get good view synthesis quality. Moreover, while training view augmentation doesn't necessarily provide better interpolated view synthesis quality, it shows significantly better results on extrapolated views.

\section{Mesh Extraction \& Scene Integration}
\label{sec:supp-mesh}

Since our primary goal is to support navigation and collision checking rather than fine-grained physical interaction, we design the mesh representation to be occupancy-accurate and efficient rather than visually detailed. Starting from the global metric point cloud, we first filter out points that are too far from the sensor trajectories, as these regions are unlikely to be reachable or relevant for agent interaction. We then voxelize the remaining points into a 3D occupancy grid and run a standard Marching Cubes~\cite{loresen1987marching} algorithm to extract a triangle mesh. This occupancy-based reconstruction provides a controllable trade-off between resolution and complexity while preserving the spatial support of walkable surfaces and major obstacles. After meshing, we perform a lightweight cleaning step to remove residual noise and irrelevant details. We discard small isolated components (\textit{e.g.}, mesh fragments with fewer than 50 faces), which typically correspond to transient objects or reconstruction artifacts. The resulting mesh (\cref{fig:supp-mesh}) does not need to be strictly watertight, as visual appearance is handled by the 3DGS model. 

We integrate the learned 3DGS model and the collision mesh into a single USD scene, using the MetaCam world coordinate frame (in meters) as the common reference. In this representation, the mesh provides a lightweight physics and collision layer, while the 3DGS model is used as the primary renderer. The resulting USD scenes can be directly loaded into simulators such as Isaac Sim, and the same collision mesh can be imported into Unity for baking navigation meshes as described below.

\section{Expert Trajectory and VLN}
\label{sec:supp-vln}
 
\textbf{Expert trajectory}. We import the collision mesh into Unity and use its built-in NavMesh baking API to extract a triangulated navigable surface. Given any pair of start and goal locations, we align them onto the closest points on the NavMesh and invoke Unity's pathfinding module to compute a collision-free expert trajectory on this surface. To ensure that the paths are semantically meaningful and well grounded in the captured data, start and goal candidates are sampled in the vicinity of camera poses from the training or test splits, so that each endpoint corresponds to a visually interpretable location in the scene.

\noindent \textbf{Language instruction}. Once an expert trajectory is obtained, we replay it in our 3DGS-based simulator to render an egocentric video along the path. We then feed both the rendered video and the corresponding trajectory summary to the Gemini 2.5 Flash model~\cite{comanici2025gemini} to automatically generate natural-language instructions. To improve controllability and consistency, we adopt a two-stage prompting scheme: the model is first asked to output a structured JSON description that segments the route into sub-instructions with associated landmarks and actions, and is then prompted to condense this JSON representation into a single fluent instruction. Compared to fully manual annotation, this procedure is substantially more scalable and yields more stable instructions across scenes, while still allowing for quality control: we perform manual spot checks and discard the very small fraction of instructions that are clearly inconsistent or unusable. The resulting set of expert trajectories and instructions is stored together with the underlying 3DGS scenes, making Wanderland a unified testbed for different embodied navigation tasks.

\section{More Experiment Details}
\subsection{3D Reconstruction} \label{sec:supp-3d}

\textbf{Evaluation metrics}. We use different evaluation metrics to assess different aspects in camera pose estimation accuracy. The definitions are detailed below:

\begin{itemize}
    \item Absolute Trajectory Error (ATE) measures the global consistency between predicted and ground truth trajectories after alignment. 
    
    \item Translation ATE - Raw (\textbf{T-ATE\textsuperscript{R}}): Root mean square error (RMSE) of translation after SE(3) alignment:
    \begin{equation}
    \text{T-ATE\textsuperscript{R}} = \sqrt{\frac{1}{N} \sum_{i=1}^N \|(R_{SE3} \cdot \mathbf{t}_i^{pred} + \mathbf{t}_{SE3}) - \mathbf{t}_i^{gt}\|^2},
    \end{equation}
    where $R_{SE3}$ and $\mathbf{t}_{SE3}$ are the rotation and translation from SE(3) alignment.
    
    \item Translation ATE - Scaled (\textbf{T-ATE\textsuperscript{S}}):
    RMSE of translation after SIM(3) alignment. This metric evaluates relative-scale pose accuracy independent of absolute scale:
    \begin{equation}
    \text{T-ATE}^{\text{S}} =
    \sqrt{\frac{1}{N} \sum_{i=1}^N
    \bigl\| s \cdot R_{\mathrm{SIM3}} \mathbf{t}_i^{\mathrm{pred}}
    + \mathbf{t}_{\mathrm{SIM3}} - \mathbf{t}_i^{\mathrm{gt}} \bigr\|^2 } ,
    \end{equation}
    where $R_{SIM3}$, $\mathbf{t}_{SIM3}$, and $s$ are the rotation, translation, and scale from SIM(3) alignment.
    
    \item Rotation ATE (\textbf{R-ATE}): RMSE of rotation angles:
    \begin{equation}
    \begin{split}
    & \text{R-ATE} = \sqrt{\frac{1}{N} \sum_{i=1}^N \Delta\theta_i^2}, \\ & \Delta\theta_i = \cos^{-1}\left(\frac{\text{tr}(\mathbf{R}_i^{gt\top} \mathbf{R}_i^{pred}) - 1}{2}\right) ,
    \end{split}
    \end{equation}
    where $\mathbf{R}_i$ denotes the rotation matrix.
    
    \item Relative Trajectory Error (RTE) measures the consistency of relative motions between camera pairs.

    \item Translation RTE (\textbf{T-RTE}):
    RMSE of relative translation distances between all camera pairs:
    \begin{equation}
    \begin{split}
     & \text{T-RTE}
    = \sqrt{\frac{2}{N(N-1)}
    \sum_{i=1}^{N-1} \sum_{j=i+1}^N
    \left( \Delta_{ij} \right)^2 } ,
    \\
     & \Delta_{ij}
    = \|\mathbf{t}_i^{pred} - \mathbf{t}_j^{pred}\|
     - \|\mathbf{t}_i^{gt} - \mathbf{t}_j^{gt}\| .
    \end{split}
    \end{equation}

    \item \textbf{T-RTE (degrees)}: RMSE of angular differences in relative translation directions. Similarly to T-ATE\textsuperscript{S}, it also evaluates relative-scale pose accuracy independent of absolute scale:
    \begin{equation}
    \begin{split}
     & \text{T-RTE (deg)} = \sqrt{\frac{2}{N(N-1)} \sum_{i=1}^{N-1} \sum_{j=i+1}^N \theta_{ij}^2}, \\
     & \quad \theta_{ij} = \cos^{-1}\left(\frac{(\mathbf{t}_i^{pred} - \mathbf{t}_j^{pred}) \cdot (\mathbf{t}_i^{gt} - \mathbf{t}_j^{gt})}{\|\mathbf{t}_i^{pred} - \mathbf{t}_j^{pred}\| \|\mathbf{t}_i^{gt} - \mathbf{t}_j^{gt}\|}\right).
    \end{split}
    \end{equation}

    \item Rotation RTE (\textbf{R-RTE}): RMSE of relative rotation angles between all camera pairs:
    \begin{equation}
    \begin{split}
     & \text{R-RTE} = \sqrt{\frac{2}{N(N-1)} \sum_{i=1}^{N-1} \sum_{j=i+1}^N \Delta\theta_{ij}^2}, \\
     & \quad \Delta\theta_{ij} = \cos^{-1}\left(\frac{\text{tr}(\mathbf{R}_{ij}^{gt\top} \mathbf{R}_{ij}^{pred}) - 1}{2}\right),
    \end{split}
    \end{equation}
    where $\mathbf{R}_{ij} = \mathbf{R}_i\mathbf{R}_j^\top$ represents the relative rotation.

    \item \textbf{AUC@30:} Area under the curve of the cumulative distribution of maximum relative errors, with a maximum threshold of 30 degrees:
    \begin{equation}
    \text{AUC@30} = \int_{0}^{30} P(\max(\text{R-RTE}, \text{T-RTE}_{\text{deg}}) < \theta)  d\theta.
    \end{equation}
    This provides a comprehensive measure of reconstruction quality across different error tolerances.
    \item Success Rate (\textbf{SR}) is defined to be the ratio of scenes with AUC@30 $> 0.1$.
\end{itemize}

\noindent \textbf{Evaluation setup}. Due to GPU memory limitation, some reconstruction models (DUSt3R and VGGT) can't process all images in a scene. To ensure fair comparison, we uniformly downsample all scenes to be below 500 images. This is another limitation of these methods.

\begin{figure*}
    \centering
    \includegraphics[width=0.88\linewidth]{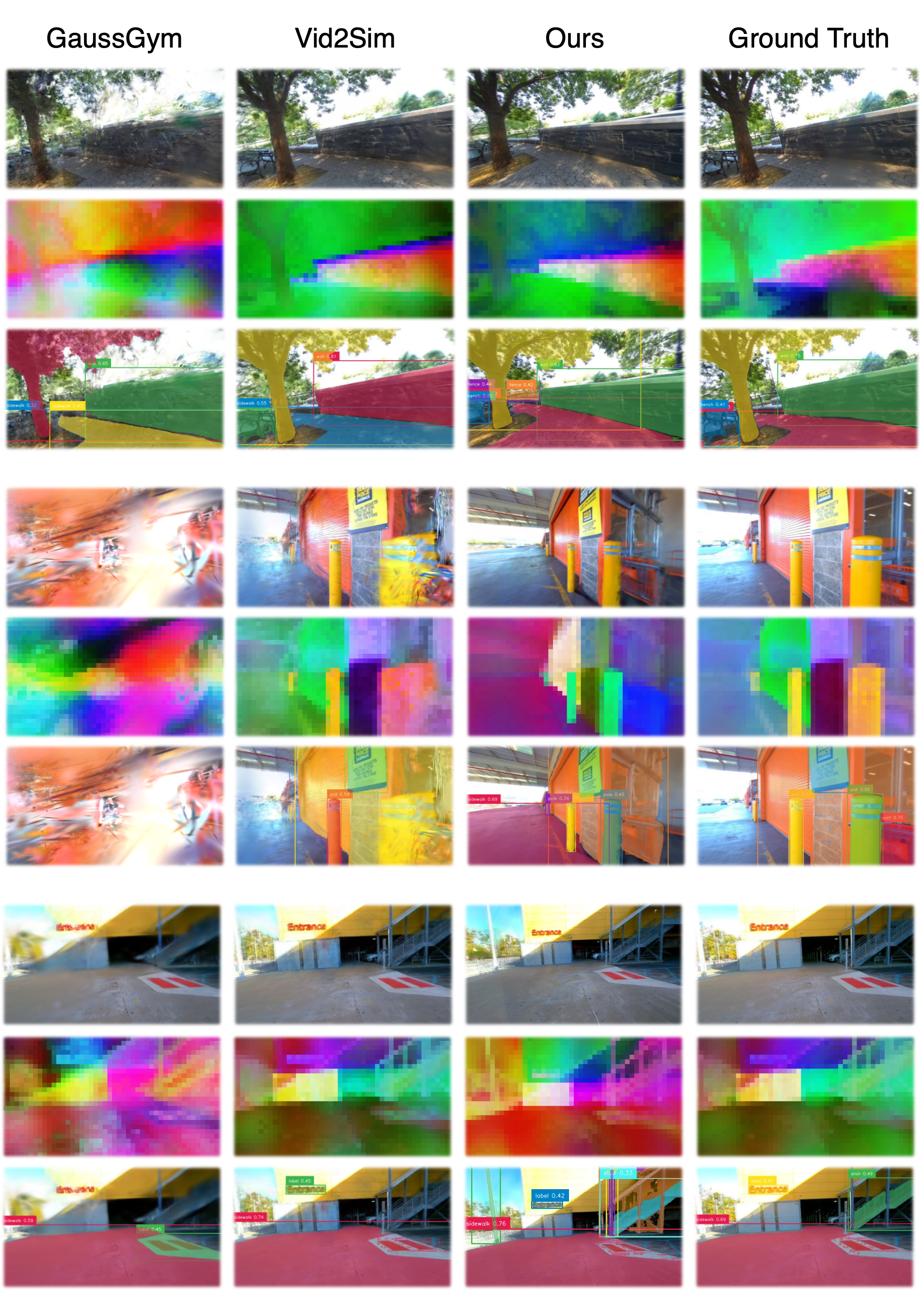}
    \vspace{-2mm}
    \caption{\textbf{More visualization on photorealistic and semantic consistent sensor simulation}. Format is the same as \cref{fig:nvs}.}
    \label{fig:supp-nvs}
\end{figure*}

\subsection{Photorealistic Sensor Simulation} \label{sec:supp-nvs}
\textbf{Evaluation metrics}. We use common metrics for evaluating novel view synthesis models:
\begin{itemize}
    \item Peak Signal-to-Noise Ratio (\textbf{PSNR}): 
    Measures pixel-wise reconstruction quality:
    \begin{equation}
    \text{PSNR} = 10 \log_{10}\left(\frac{1}{\text{MSE}}\right),
    \end{equation}
    where $\text{MSE} = \frac{1}{WH}\sum_{i,j}(I^{pred}_{i,j} - I^{gt}_{i,j})^2$.

    \item Structural Similarity Index (\textbf{SSIM}): 
    Evaluates structural preservation:
    \begin{equation}
    \text{SSIM} = \frac{(2\mu_x\mu_y + C_1)(2\sigma_{xy} + C_2)}{(\mu_x^2 + \mu_y^2 + C_1)(\sigma_x^2 + \sigma_y^2 + C_2)},
    \end{equation}
    where $\mu,\sigma$ are local statistics.

    \item Learned Perceptual Image Patch Similarity (\textbf{LPIPS}): 
    Measures perceptual quality using deep features:
    \begin{equation}
    \text{LPIPS} = \sum_l \frac{1}{H_lW_l} \sum_{h,w} \| w_l \odot (\phi_l(I^{pred})_{h,w} - \phi_l(I^{gt})_{h,w}) \|_2^2,
    \end{equation}
    where \(\phi_l\) is the deep features from the VGGNet~\cite{simonyan2014very}.
\end{itemize}

\noindent \textbf{More results}. \Cref{fig:supp-nvs} shows more qualitative comparisons on different sim-to-real pipelines.

\subsection{Embodied Navigation} \label{sec:supp-navigation}
\textbf{Evaluation metrics}. We use three common metrics and a self-designed metric to evaluate embodied navigation tasks:
\begin{itemize}
    \item Navigation Error (\textbf{NE}): Euclidean distance between the agent's final position and the goal location upon episode termination.

    \item Success Rate (\textbf{SR}): Percentage of episodes where the agent successfully reaches the goal within the specified maximum steps.
    
    \item Success weighted by Path Length (\textbf{SPL}): Combined metric considering both success and path efficiency:
    \begin{equation}
    \text{SPL} = \frac{1}{N} \sum_{i=1}^N S_i \frac{l_i}{\max(p_i, l_i)},
    \end{equation}
    where $S_i$ is success (0/1), $l_i$ is optimal path length, and $p_i$ is actual path length for episode $i$.
    
    \item Intervention Rate (\textbf{IR}): Percentage of episodes requiring early termination due to non-timeout failures including agent becoming stuck, making no progress, or exiting scene boundaries.
\end{itemize}
\paragraph{RL training setup.}
For the ``RL post-trained'' rows in our experiments, we fine-tune the released navigation policies of NoMaD, CityWalker, and MBRA on Wanderland using a standard on-policy reinforcement learning setup. 
All agents interact with our 3DGS-based simulator described in the main paper and operate in the same action space as in their original implementations. 
Episodes are sampled from the training split by drawing start and goal locations on the navigable NavMesh as in \cref{sec:wanderland}, and each episode is capped at a fixed maximum number of steps (1000 in our experiments). 
Episodes may also terminate early when the agent reaches the goal, gets stuck, or leaves the valid navigation region. 
We use Proximal Policy Optimization (PPO) with generalized advantage estimation (GAE), a $\gamma=0.99$ discount factor.
Unless otherwise noted, we only update the policy, so that RL mainly adapts high-level navigation behavior to the geometry and semantics of Wanderland.

\paragraph{Reward design.}
The reward function follows a simple distance-based shaping scheme with explicit penalties for unsafe behaviors. 
Let $d_t$ denote the distance from the agent to the goal at time step $t$. 
The per-step reward $r_t$ is defined as
\begin{equation}
r_t =
\begin{cases}
+R_{\text{succ}}, & \text{if the agent reaches the goal}, \\[3pt]
-R_{\text{fail}}, & \text{if the episode early terminated}, \\[3pt]
-\alpha + \beta \,(d_{t-1} - d_t), & \text{otherwise},
\end{cases}
\end{equation}
where $R_{\text{fail}} > 0$ is a terminal success bonus, $R_{\text{fail}} > 0$ controls the penalty for unsafe terminations, $\alpha > 0$ is a small step penalty that encourages shorter paths, and $\beta > 0$ weights the progress reward given by the reduction in geodesic distance. 
In words, the agent is rewarded for making progress toward the goal, slightly penalized for every time step, strongly rewarded upon success, and explicitly penalized when it falls, gets stuck, or ignores obstacles and exits the valid navigation area. 
This shaping aligns the optimization objective with our evaluation metrics: minimizing NE by rewarding geodesic progress, maximizing SR by giving success bonuses, improving SPL by discouraging unnecessarily long paths, and reducing IR by penalizing unsafe behaviors that trigger early termination. 
We use a single set of reward coefficients across all methods to ensure a fair comparison of RL post-training effects on Wanderland.


\end{document}